\documentclass[10pt,journal,compsoc]{IEEEtran}

\usepackage{float}
\usepackage{color}
\usepackage[english]{babel}
\usepackage{multirow}
\usepackage{makecell}
\usepackage{graphicx}
\usepackage{url}
\usepackage{subfigure}
\usepackage{dsfont}
\usepackage{hyperref}
\usepackage{booktabs}
\usepackage{amsmath}
\usepackage{amsfonts}
\usepackage{array}
\usepackage{setspace}
\usepackage{arydshln}
\usepackage{adjustbox}
\usepackage{ifthen}
\usepackage{bm}

\newcommand{\name}{4Dynamic}

\ifCLASSOPTIONcompsoc
  \usepackage[nocompress]{cite}
\else
  \usepackage{cite}
\fi

\begin{document}
\title{\name: Text-to-4D Generation \\ with Hybrid Priors}

\author{Yu-Jie Yuan, Leif Kobbelt, Jiwen Liu, Yuan Zhang, Pengfei Wan, Yu-Kun Lai, and Lin~Gao\IEEEauthorrefmark{1}
	\thanks{\IEEEauthorrefmark{1} Corresponding Author is Lin Gao (gaolin@ict.ac.cn).}}

\markboth{IEEE Transactions on Pattern Analysis and Machine Intelligence,~Vol.~xx, No.~xx, April~2024}
{\name}

\IEEEtitleabstractindextext{%
\begin{abstract}
Due to the fascinating generative performance of text-to-image diffusion models, growing text-to-3D generation works explore distilling the 2D generative priors into 3D, using the score distillation sampling (SDS) loss, to bypass the data scarcity problem. The existing text-to-3D methods have achieved promising results in realism and 3D consistency, but text-to-4D generation still faces challenges, including lack of realism and insufficient dynamic motions. In this paper, we propose a novel method for text-to-4D generation, which ensures the dynamic amplitude and authenticity through direct supervision provided by a video prior. Specifically, we adopt a text-to-video diffusion model to generate a reference video and divide 4D generation into two stages: static generation and dynamic generation. The static 3D generation is achieved under the guidance of the input text and the first frame of the reference video, while in the dynamic generation stage, we introduce a customized SDS loss to ensure multi-view consistency, a video-based SDS loss to %
improve
temporal consistency, and most importantly, direct priors from the reference video to ensure the quality of geometry and texture. Moreover, we design a prior-switching training strategy to avoid conflicts between different priors and fully leverage the benefits of each prior. In addition, to enrich the generated motion, we further introduce a dynamic modeling representation composed of a deformation network and a topology network, which ensures dynamic continuity while modeling topological changes. Our method not only supports text-to-4D generation but also enables 4D generation from monocular videos. The comparison experiments demonstrate the superiority of our method compared to existing methods.
\end{abstract}

}

\maketitle

\IEEEdisplaynontitleabstractindextext

\IEEEpeerreviewmaketitle

\IEEEraisesectionheading{\section{Introduction}}

\begin{figure*}[!t]
    \centering
    \includegraphics[width=0.97\linewidth]{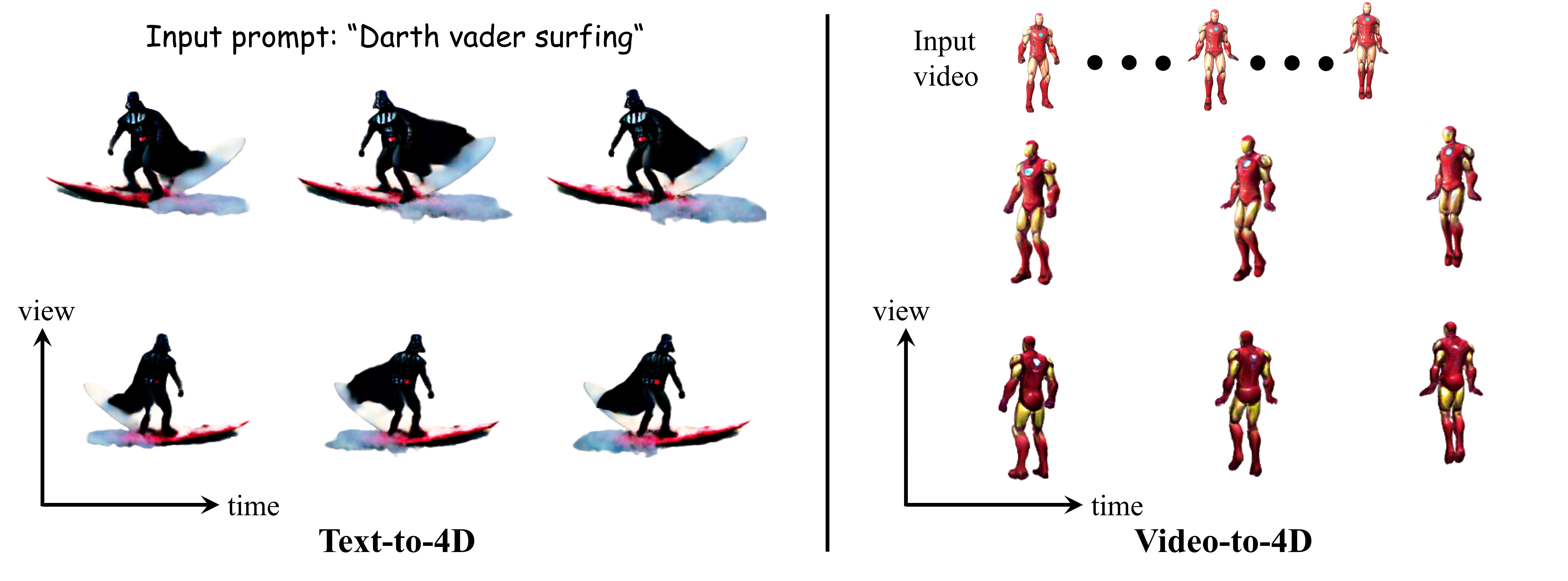}
    \vspace{-5mm}
    \caption{%
    We propose a novel text-to-4D NeRF generation method, \name, which exploits not only a generative prior from the score distillation sampling (SDS) but also a direct prior from the pre-generated reference video. As a result, our method can achieve high-quality 4D generation from a text prompt or a monocular video. }
    \vspace{-6mm}
    \label{fig:teaser}
\end{figure*}

\IEEEPARstart{T}{he} production and generation of digital content (both 2D and 3D) traditionally rely on capturing from the real world or manual creation by professionals which is cumbersome and expensive. With the prosperity of high-quality large-scale generative models, especially cross-modal models such as those based on diffusion models, 2D images can be generated from various text prompts. This has brought about the popularity of generative artificial intelligence, which makes professional content generation (PCG) converge to user content generation (UCG), allowing casual users to create diverse imaginary 2D content. 
For 3D models, existing generative models are often custom designed and trained individually for different representations, such as voxels~\cite{3dgan}, point clouds~\cite{spgn}, or meshes~\cite{SDM-NET,yang2022dsg}, and also rely on a large scale of high-quality 3D datasets~\cite{deitke2023objaverse,deitke2023objaversexl} and can only generate novel shapes on limited categories. Implicit representations, such as Neural Radiance Fields (NeRF)~\cite{mildenhall2020nerf}, are gradually replacing explicit representations as the core representation in 3D generative models. With the help of tri-planes~\cite{eg3d} or voxels~\cite{liu2020neural, mueller2022instant}, it is convenient to introduce 2D generative networks, such as StyleGAN~\cite{karras2019style} or diffusion models~\cite{rombach2022high} to generate 3D NeRF~\cite{eg3d,gao2022get3d,wang2023rodin,chen2023single,karnewar2023holodiffusion,karnewar2023holofusion}. However, the generation quality still depends on the quantity and quality of the training data.

Due to the lack of high-quality 3D data and inefficiency in manual collection and creation, exploiting existing large-scale generative models to assist in 3D generation has become a viable approach. The text-to-image diffusion model is a powerful tool, which is trained on a large-scale paired text-image dataset~\cite{schuhmann2022laion} and has fascinating generation performance. Therefore, DreamFusion~\cite{poole2022dreamfusion}, as a pioneering work, combines it with NeRF representation and proposes the score distillation sampling (SDS) loss. Guided by the input text, a pre-trained diffusion model is used to supervise the rendered images from NeRF, achieving text-based 3D NeRF generation. This work has sparked a series of works around this formulation and the field is developing rapidly. Some remarkable text-to-3D or image-to-3D results have been achieved by introducing 3D priors to enhance 3D consistency~\cite{liu2023zero,shi2023mvdream}, improving SDS loss to some variants~\cite{wang2023prolificdreamer,yu2023text}
or introducing physical-based materials based on mesh rendering~\cite{chen2023fantasia3d} to improve texture synthesis quality.

However, the generated objects are all static, even though motion is an inherent property of our world. It is a laborious task to endow generated 3D objects with motion, and users %
often prefer
to use simple inputs, such as text prompt. Therefore, generating a dynamic scene from a text similar to existing 3D generation methods remains to be a problem that needs to be %
addressed.
The methods for reconstructing a dynamic NeRF are emerging one after another, but generating a dynamic NeRF from a text prompt still poses great challenges. The main challenge is the augmentation in data dimensions between input and output. The differences in dimensions need to be compensated for by introducing other information, such as the priors from pre-trained generative models. However, existing data does not support training a large model for 4D generation, so current methods use SDS to distill from a 2D generative model. 
The pioneering work, MAV3D~\cite{singer2023text}, adopts HexPlane~\cite{cao2023hexplane} as the dynamic representation and divides the generation into three stages: static, dynamic, and super-resolution with a combination of image SDS loss and video SDS loss. The latter one distills 4D dynamic information from the text-to-video diffusion model. However, the generation results are not satisfactory.

In this paper, we propose a novel text-to-4D generation pipeline, which not only introduces a video SDS loss based on the existing text-to-video diffusion model but also obtains additional supervision from the pre-generated reference video to improve the quality of 4D generation. The input text will first generate a reference video with the help of a text-to-video generative model and the generated video will be used as a direct prior for our 4D generation. Then the remaining generation process is divided into two stages: static 3D generation and dynamic generation. The static stage adopts joint supervision from 2D SDS and 3D SDS losses to ensure diversity and 3D consistency. The first frame of the generated video will be used as the reference image to ensure the alignment between the generated result and the generated video prior. 
In the following dynamic stage, we introduce a dynamic representation composed of a deformation network and a topology network which model the continuous motion as well as topological changes, respectively, ensuring the continuity and diversity 
of the generated motion. In terms of supervision, we also introduce a video SDS loss distilling the generative prior from the 2D text-to-video diffusion model. More importantly, we exploit additional supervision from the reference video. We first build a customized SDS loss based on the result of the static stage. The video SDS loss mainly ensures the temporal consistency, while the customized SDS loss ensures the 3D consistency of geometry and texture. However, the supervision of SDS loss is not direct, so we further incorporate direct losses from the reference video prior. We adopt multiple losses from different priors in the dynamic generation stage, namely the direct prior from the reference video and the generative prior from the diffusion models, so we design a prior-switching training strategy. Specifically, in the early iterations of training, we rely on the direct prior to guide and stabilize motion generation, while in the later iterations, we gradually transition to the distillation of the diffusion model to enhance motion amplitude and diversity.
The contributions of our method are summarized as follows:

\begin{itemize}
    \item We introduce a dynamic representation composed of a deformation field and a topology field in text-to-4D generation. The former ensures continuous and sufficient dynamics, while the latter is responsible for topologically discontinuous changes, helping achieve diverse dynamic generation.
    \item We propose a novel text-to-4D generation method with a prior-switching training strategy that not only exploits the text-to-video diffusion model for SDS supervision but also builds additional supervision from the pre-generated reference video to ensure generation quality and dynamic effects.
    \item Our method can achieve not only text-to-4D generation but also 4D generation from monocular videos. It achieves state-of-the-art 4D generation performance compared to existing methods.
\end{itemize}

\vspace{-4mm}

\section{Related Work}

\subsection{Dynamic Neural Radiance Fields}
Neural Radiance Field (NeRF)~\cite{mildenhall2020nerf} has become a popular 3D implicit representation in recent times. Due to its fascinating results in novel view synthesis and 3D reconstruction~\cite{wang2021neus}, it has been further extended for digital human modeling~\cite{peng2021neural,zhang2021editable,liu2021neural,weng2022humannerf,jiang2022neuman,liu2023hosnerf}, better rendering effects~\cite{barron2021mip,hu2023tri}, generalization on different scenes~\cite{wang2021ibrnet,chen2021mvsnerf}, faster training or inference speed~\cite{garbin2021fastnerf,yu2021plenoctrees,hedman2021baking,mueller2022instant,Chen2022tensorf,kerbl20233d}, 
geometry or appearance editing~\cite{liu2021editing,yuan2022nerf,huang2022stylizednerf,neumesh,NerfFaceEditing}, etc.
For more comprehensive and detailed discussions and comparisons, we refer the readers to these surveys~\cite{dellaert2020neural,gao2022nerf,tewari2022advances}.

Our work focuses on the text-based generation of 4D scenes represented by dynamic NeRF~\cite{pumarola2021d,tretschk2021non}, which adds additional temporal inputs to NeRF. 
Some works directly take the positionally encoded time~\cite{li2021nsff,xian2021space} or learnable vectors~\cite{li2021neural} as one of the NeRF inputs and encode spatial and temporal information in a single network simultaneously. They typically introduce additional supervision, such as the predicted depth~\cite{xian2021space} and cycle consistency of scene flow~\cite{li2021nsff,du2021neural}. This kind of method can be further improved by utilizing the discrete cosine transform representation of scene flow~\cite{wang2021neural} or separately modeling different scene parts~\cite{gao2021dynamic,wang2022mixed,song2022nerfplayer}.
Another kind of method predicts the offset~\cite{pumarola2021d,tretschk2021non} or SE(3) transformation field~\cite{park2021nerfies} for each sampled point by an additional network. The elastic energy constraint is introduced to constrain the Jacobian matrix of the transformation~\cite{park2021nerfies,tretschk2021non}. 
Further, to handle topological changes, HyperNeRF~\cite{park2021hypernerf} regards different topological states as hyperplanes of a high-dimensional space and introduces a topology network. 
With the emergence of NeRF acceleration methods~\cite{mueller2022instant}, dynamic NeRF is also accelerated by introducing an explicit voxel representation~\cite{li2022streaming} or a combination of tri-planes/voxels and implicit networks~\cite{shao2022Tensor4d,jang2022d,fang2022fast,kappel2022fast,cao2023hexplane}.
Our method adopts a dynamic modeling approach that combines the deformation and topology networks, where the former ensures dynamic continuity and amplitude, and the latter models potential topological changes, which is flexible.

\subsection{Text/Image-guided 3D Generation}
Although one can also generate a 3D object from the given text through training a specific generative model to generate a tri-plane representation of NeRF~\cite{wang2023rodin,li2023instant3d}, this way is limited by the size and quality of the 3D dataset. With the introduction of the SDS (Score Distillation Sampling) loss~\cite{poole2022dreamfusion}, utilizing a 2D pre-trained diffusion model~\cite{rombach2022high} to distill the generative power to text-based 3D generation becomes a popular solution. However, realism and 3D consistency are two main facing challenges.
Magic3D~\cite{lin2023magic3d} extends DreamFusion by performing secondary optimization on the extracted mesh with mesh-based differentiable rendering to render high-resolution images. SJC~\cite{wang2023score} proposes a variant of SDS while multiple improved versions of SDS are also proposed~\cite{wang2023prolificdreamer,yu2023text,zhu2023hifa,liang2023luciddreamer}. Fantasia3D~\cite{chen2023fantasia3d} divides the 3D generation into geometry and texture stages and introduces physical-based material as the texture representation, further enhancing the appearance realism of the generation. DreamTime~\cite{Huang2023DreamTimeAI} improves the generation quality by modifying the timestep sampling strategy. DreamBooth3D~\cite{Raj2023DreamBooth3DST} achieves customized generation through the use of DreamBooth~\cite{Ruiz2022DreamBoothFT}.
The ``multi-face'' or ``Janus'' problem is the dimension curse when using 2D diffusion models for 3D generation. So the 3D prior is introduced to solve this curse. The 3D shape can be directly given~\cite{metzer2023latent} or estimated from the image~\cite{xu2023dream3d}, providing geometric initial values for optimizing NeRF. MVDream~\cite{shi2023mvdream} proposes to fine-tune the diffusion model to generate multi-view images and as so explicitly embeds 3D information into a 2D diffusion model. Using the fine-tuned model to generate 3D NeRF effectively alleviates the Janus problem.
Based on these text-to-3D methods, we can achieve image-to-3D with the image as an additional input~\cite{deng2023nerdi,gu2023nerfdiff}. Magic123~\cite{qian2023magic123} and Make-it-3D~\cite{tang2023make} add additional supervision using the input image during the SDS optimization, while Zero123~\cite{liu2023zero} fine-tunes the diffusion model by changing the condition to the image and the relative view. The follow-up works not only improve the synthesis quality~\cite{ye2023consistent,liu2023syncdreamer,zeng2023ipdreamer,yu2023hifi,shi2023zero123++,weng2023consistent123,lin2023consistent123,sun2023dreamcraft3d,long2023wonder3d,Wang2023ImageDreamIM} but also consider how to increase efficiency~\cite{liu2023one,liu2023one++,hong2023lrm}.
By utilizing text-based generation capabilities, text-based editing can also be achieved~\cite{haque2023instruct,sella2023vox,zhuang2023dreameditor,li2023focaldreamer,cheng2023progressive3d,mirzaei2023watch,fang2023text}.
Due to the rise of 3D Gaussian Splatting (3DGS)~\cite{kerbl20233d}, some methods~\cite{Tang2023DreamGaussianGG,Yi2023GaussianDreamerFG,Chen2023Textto3DUG} have replaced NeRF representation with 3DGS to achieve multi-view generation.
For a summary of these methods, please refer to this survey~\cite{li2023generative}.

\begin{figure*}[!t]
    \centering
    \includegraphics[width=0.97\linewidth]{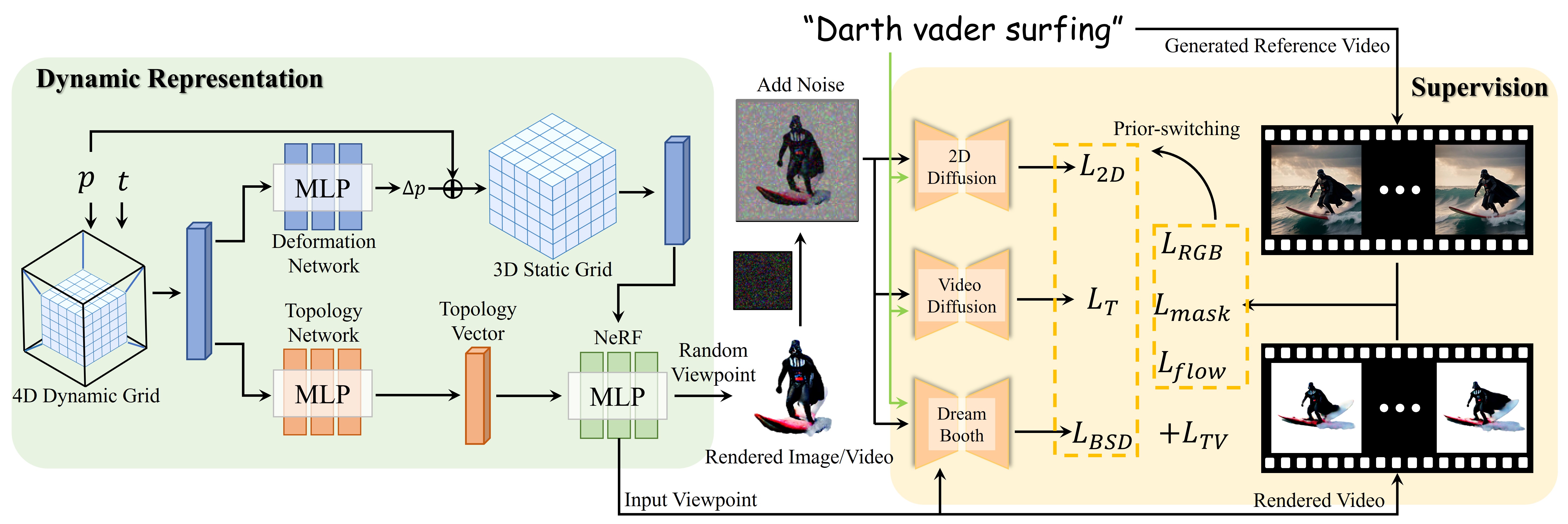}
    \vspace{-4mm}
    \caption{The pipeline of our method. We mainly show the dynamic generation process here. A dynamic representation consisting of a deformation network and a topology network is introduced. The NeRF network includes a density network and a color network. The rendered image or video under a random viewpoint is supervised by 2D SDS, customized SDS (BSD) and video SDS losses. Moreover, we exploit the pre-generated reference video to provide direct supervision under the input viewpoint. To balance between different priors, we design a prior-switching training strategy to achieve generation results that have good dynamic motion.}
    \vspace{-4mm}
    \label{fig:pipeline}
\end{figure*}

\subsection{Text/Video-guided 4D Generation}
Our method focuses on 4D NeRF generation from text, which is more challenging than those text/image-based 3D generation methods mentioned above. The pioneering work, MAV3D~\cite{singer2023text} introduces video-based SDS loss and adopts dynamic NeRF representation, HexPlane~\cite{cao2023hexplane}. It divides the generation process into three stages: static, dynamic, and super-resolution, but the generation quality can be further improved. 
Recently, Dream-in-4D~\cite{zheng2023unified} adopts a dynamic NeRF representation based on the deformation field and divides the text-to-4D generation into static and dynamic stages. 4D-fy~\cite{bahmani20234d} adopts a hybrid feature representation of dynamic and static voxels and proposes a hybrid optimization strategy with the combination of SDS loss~\cite{poole2022dreamfusion}, 3D SDS loss~\cite{shi2023mvdream} and video SDS loss~\cite{Luo2023VideoFusionDD}. By assigning a dynamic network on 3DGS and optimizing it under the video SDS and the constraint on 3D Gaussians, dynamic generation can also be achieved~\cite{Ling2023AlignYG}.
Based on 4D-fy, TC4D~\cite{bahmani2024tc4d} decomposes the motion into the global motion parameterized by a spline curve and the local motion of each object itself. The former is a motion trajectory specified by the user, while the latter is generated segment-by-segment with the video-based SDS loss.
Comp4D~\cite{xu2024comp4d} splits the input prompt into different entities
using a Large Language Model (LLM), generates 4D objects separately, and then combines them using the trajectory information given by the LLM.
Another solution to text-based 4D generation is 4D reconstruction from a monocular video, rather than using SDS loss. For example, Vidu4D~\cite{wang2024vidu4d} generates a video from the text prompt and then employs dynamic Gaussian surfels for 4D reconstruction, while Diffusion4D~\cite{liang2024diffusion4d} fine-tunes the video diffusion model to generate the orbital video for 4D reconstruction.
EG4D~\cite{sun2024eg4d} first adopts attention injection to generate consistent multi-view videos, and then, after 4D Gaussian Splatting (4D-GS)~\cite{wu20244d}
reconstruction, fine-tuning is conducted using a diffusion model prior.
PLA4D~\cite{miao2024pla4d} aligns 3DGS to the mesh generated by an image-to-mesh generative model and generates dynamics through the pixel loss and SDS supervision provided by Zero123~\cite{liu2023zero}, with static images serving as references.
Some works employ the Material Point Method in physical simulation to achieve motion transfer~\cite{fu2024sync4d} and interactive dynamic generation~\cite{zhang2024physdreamer} based on a monocular video.
In addition to generating 4D NeRF from text, there are also Consistent4D~\cite{jiang2023consistent4d} and 4DGen~\cite{yin20234dgen} that generate 4D content from a monocular video and Animate124~\cite{zhao2023animate124} that generates 4D NeRF from an image with a text.
AnimatableDreamer~\cite{Wang2023AnimatableDreamerTN} and MagicPose4D~\cite{zhang2024magicpose4d} explore the 4D reconstruction and motion transfer of articulated objects from monocular videos.
SC4D~\cite{wu2024sc4d} adopts the dynamic 3DGS of SC-GS (Sparse Controlled Gaussian Splatting)~\cite{huang2023sc}, and achieves video-to-4D generation through coarse-to-fine optimization.
DreamScene4D~\cite{chu2024dreamscene4d} can reconstruct 4D scenes from in-the-wild videos containing multiple objects. It segments, tracks, and reconstructs different objects separately, and combines them with background using monocular depth prediction guidance.
STAG4D~\cite{zeng2024stag4d} generates 6 additional reference videos from input or generated videos through a spatial and temporal attention fusion mechanism and selects adjacent-view images as references in multi-view SDS loss.
4Diffusion~\cite{zhang20244diffusion} fine-tunes ImageDream~\cite{Wang2023ImageDreamIM} to incorporate temporal consistency, and then utilizes the fine-tuned model for SDS loss while adopting image-level perceptual loss for supervision.

The above methods either adopt the SDS-based optimization or utilize the video generation combined with 4D reconstruction. However, the former distills the generative prior which lacks direct supervision, and the latter suffers from inconsistencies in generating multi-view videos or conflicts when blending video and SDS supervision. So our method incorporates a reference video for supervision and introduces a prior-switching training strategy. This not only avoids the limitation of using SDS alone but also mitigates conflicts inherent in mixed supervision. Moreover, our method employs a dynamic representation with a hybrid of deformation and topology networks to ensure both dynamic continuity and diversity.

\section{Method}

Our method generates dynamic 3D scenes from the input text. The pipeline of our method is shown in Fig.~\ref{fig:pipeline}. We will first introduce our 4D NeRF representation, where the dynamic modeling part is crucial (Sec.~\ref{sec:representation}). Then we will introduce our text-to-4D process, where we not only perform two-stage generation but also introduce additional supervision from the pre-generated reference video to each stage (Sec.~\ref{sec:framework}). Finally, we will introduce the training strategy which ensures the final generation quality (Sec.~\ref{sec:training}).

\subsection{4D Representation}
\label{sec:representation}
Our text-to-4D generation method is based on existing text/image-to-3D generation methods, so it is crucial to assign dynamic information based on the 3D generation result, which requires a good dynamic representation. Our 4D representation is a dynamic NeRF method that includes three parts: the static NeRF, the deformation network, and the topology network. First, for the static NeRF, following Magic3D~\cite{lin2023magic3d}, we adopt the multi-resolution hash grid representation Instant-NGP~\cite{mueller2022instant}. Specifically, under the sampled camera view, a camera ray $\mathbf{r}$ is injected into space from the camera center and some points $\mathbf{p}_i$ are sampled on the ray. The coordinates of each sampled point will query the features $\mathbf{f}_i$ on the 3D multi-resolution feature grid through hash-index and tri-linear interpolation. The queried features are fed into a small MLP to obtain the volume density $\sigma_i$ and color values $\mathbf{c}_i$. Note that since we do not consider complex view-dependent effect generation, the view direction is not considered in the prediction of color, which is consistent with existing 3D generation methods~\cite{qian2023magic123}. Finally, the volume densities $\sigma_i$ and colors $\mathbf{c}_i$ of all sampled points $\mathbf{p}_i$ on each ray $\mathbf{r}$ are aggregated through volume rendering~\cite{kajiya1984ray}:
\begin{equation}
    \mathbf{c}_{\mathbf{r}} = \sum_{i} {\alpha_i \mathbf{c}_i \prod_{j<i} {(1-\alpha_j)} },
\end{equation}
where $\alpha_i=1-e^{(-\sigma_i\|\mathbf{p}_{i+1}-\mathbf{p}_i\|)}$, $\mathbf{c}_{\mathbf{r}}$ is the resulting pixel color corresponding to the ray $\mathbf{r}$. During the SDS optimization, we traverse each pixel for rendering to obtain the final image.

Then we add a dynamic representation on top of the static representation to support the dynamic generation. The dynamic modeling method will affect the final dynamic result. The existing dynamic generation methods either adopt a deformation field network~\cite{zheng2023unified}, which can model motion continuity well but is limited by the topological change, or they introduce additional dynamic feature inputs through temporal feature grids~\cite{singer2023text,bahmani20234d}, which can model diverse motions but lacks continuity guarantee. 
Therefore, we propose to exploit the combination of a deformation network and a topology network to ensure motion continuity while breaking the limitation of topology and enriching motion types. First, in order to provide sufficient encoding of temporal information without affecting the encoding of static position information, we introduce a 4D multi-resolution hash-encoded feature grid to embed the 4D input composed of coordinates and time and output the temporal feature at time $t$. The output temporal feature will be input into the deformation network to predict the displacement of the corresponding sampled point, and the sampled point will be transformed back from the current observation space to the canonical space where the static model is located. In addition to the deformation network, we introduce a topology network to represent the potential topological changes. It also inputs the temporal feature output by the 4D feature grid and outputs a topology vector to describe the current topology state at time $t$. After the sampled point is transformed by the deformation network, its coordinates will be encoded through the static 3D feature grid. The topology vector is then concatenated with this static positional feature vector and input into the subsequent small MLPs to predict volume density and color.

The topology network is introduced to address the limitation that the continuous deformation network cannot represent topological changes. Therefore, compared to using only pure deformation, the extended model is more flexible and thus generates more diverse dynamics. As we will illustrate later, Fig.~\ref{fig:ablation_topo} shows an example to illustrate the function of the topology network in motion reconstruction and generation. The name ‘topology’ is because the output vector of this network accounts for the discontinuities when the topology state changes.

\subsection{4D Generation with Video Prior}
\label{sec:framework}
Based on the proposed dynamic representation, we further explore how to achieve text-to-4D generation. We divide the generation into two stages, static 3D generation and dynamic generation. Similar to 3D generation, we consider using a pre-trained text-to-video diffusion model to design the video SDS loss to supervise the dynamic generation. However, this kind of supervision is indirect and may generate implausible dynamic results. Therefore, we propose to fully utilize the text-to-video diffusion model in that we first generate a corresponding reference video from the input text, and apply it as a direct prior to dynamic generation. Then, under the guidance of video prior, the two generation stages are factored into image-guided text-to-3D generation and video-guided dynamic generation.

The first stage achieves 3D generation under the text and image inputs, where the image input comes from the first frame of the reference video. We propose to use a joint SDS loss scheme including 2D and 3D SDS losses to supervise the multi-view images rendered from the static NeRF. The 2D SDS loss $L_{2D}$ is consistent with the definition in DreamFusion~\cite{poole2022dreamfusion}, but we use the open-source model Stable Diffusion~\cite{rombach2022high}. The 3D SDS loss $L_{3D}$ has a similar formulation but adopts the Stable Zero-1-to-3 model, which has the same structure as Zero-1-to-3~\cite{liu2023zero}. It takes the input image and relative view as conditional inputs and outputs the novel view image. Note that the diffusion model used in SDS loss can be replaced with other updated models. For example, 3D SDS loss can potentially use ImageDream~\cite{Wang2023ImageDreamIM}. In addition to text-based supervision, we also add RGB loss $L_{RGB}$ and mask loss $L_{mask}$ in the initial view based on the image input to ensure that the identity of the generated 3D model is consistent with the image, which is beneficial for applying video prior to the subsequent dynamic generation. Ultimately, the optimization loss $L_{static}$ for this stage is:
\begin{equation}
    L_{static} = \lambda_{2D} L_{2D} + \lambda_{3D} L_{3D} + \lambda_{RGB} L_{RGB} + \lambda_{mask} L_{mask},
\end{equation}
where $\lambda_{2D}$, $\lambda_{3D}$, $\lambda_{RGB}$, and $\lambda_{mask}$ are the weights for corresponding losses, which are set to be 0.025, 1, 1000, 100. During the optimization, 2D SDS loss ensures generation capability while 3D SDS loss ensures multi-view consistency and minimizes the impact of the Janus problem as much as possible.

\begin{figure*}[!t]
    \centering
    \includegraphics[width=0.97\linewidth]{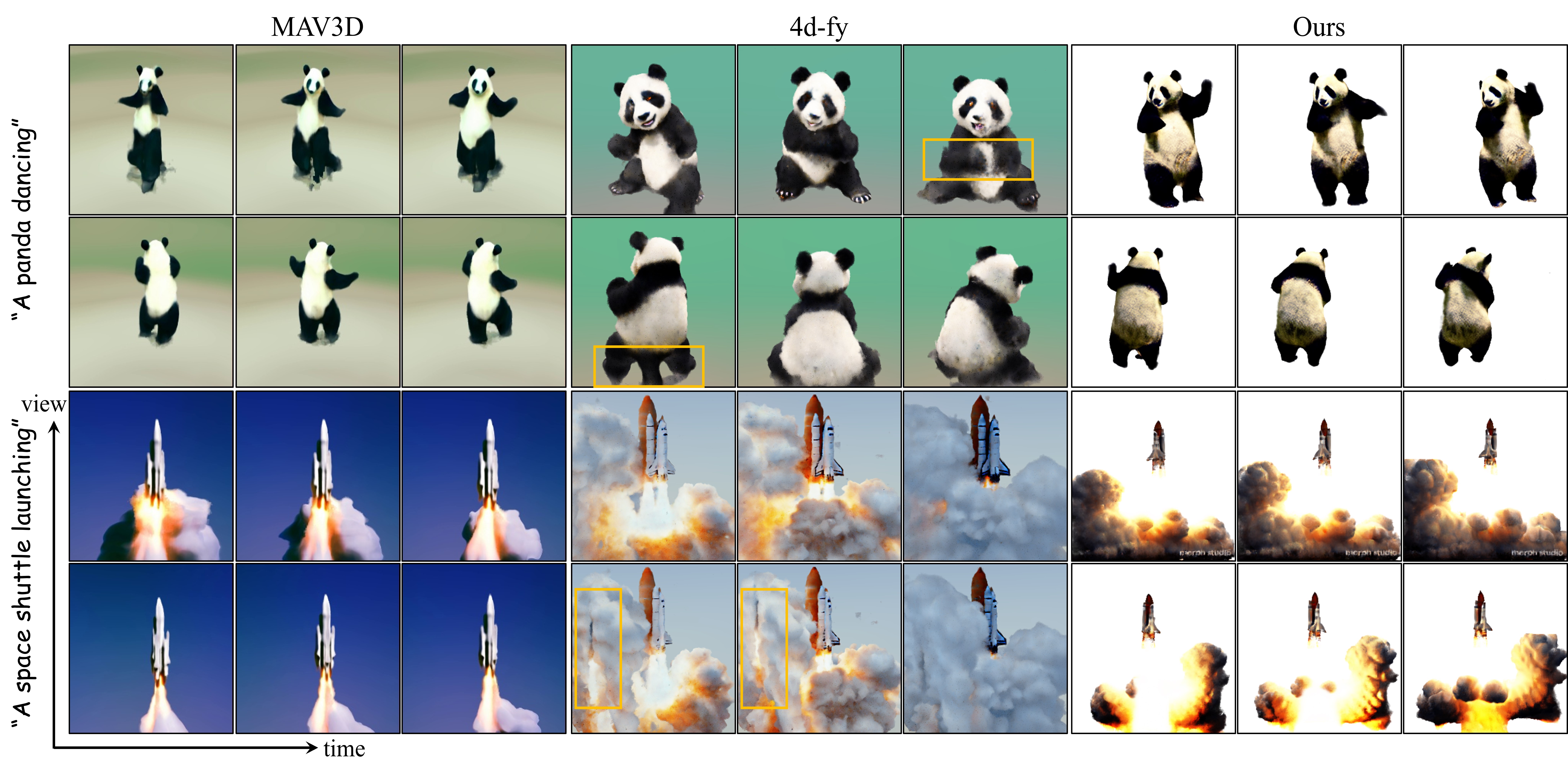}
    \vspace{-4mm}
    \caption{Comparisons of the text-to-4D generation with MAV3D~\cite{singer2023text} and 4d-fy~\cite{bahmani20234d}. Our method has significant advantages over MAV3D. For example, the panda face and the rocket have more details, and the emitted smoke looks more realistic. Compared to 4d-fy, our results are more realistic, thanks to the introduction of the direct prior from the pre-generated reference video. Some results of 4d-fy have unreasonable parts, such as a third leg growing behind the panda's back and phantoms appearing in the smoke which are marked by the orange boxes.}
    \vspace{-3mm}
    \label{fig:text_compare}
\end{figure*}

The dynamic stage incorporates the dynamic representation including the 4D feature grid, the deformation network, and the topology network into the optimization. The static part will be assigned a smaller learning rate for fine-tuning.
The dynamic generation will mainly be supervised by SDS loss based on the text-to-video diffusion model. Specifically, we sample a continuous camera trajectory and render a 24-frame video. Similar to image SDS loss, the process of adding and removing noise occurs in the latent space of the video, so video SDS loss can be defined similarly:
\begin{equation}
    L_{T}(\theta, X=f(\theta, c)) = {\mathbb{E}}_{t,c,\epsilon}[\left \| X - \hat{X}_{0} \right \|_{2}^{2} ]
\end{equation}
where $X$ is the latent representation of the rendered video from the dynamic NeRF $f(\theta, c)$ with the parameters $\theta$ and camera $c$. 
And $\hat{X}_{0}$ is the estimated $X_{0}$ based on the diffusion output $\epsilon_{\phi}(X_{t}; y,c,t)$, where $y$ is the text condition, $X_{t}$ is the diffusion forward process result at timestep $t$ with Gaussian noise $\epsilon$.
We use the Modelscope~\cite{Luo2023VideoFusionDD} video diffusion model as the 2D motion prior, and similarly, this model can be replaced with other diffusion models, such as Zeroscope~\cite{wang2023videofactory}. In addition to SDS loss, since a reference video that matches the text prompt is pre-generated, similar to the static stage, the dynamic stage can formulate direct supervision from the reference video. We assume that the reference video has a fixed camera view (which can be ensured by adding corresponding text prompts during generation) so that the corresponding video can be rendered from our 4D representation at the initial view for supervision. We introduce RGB loss $L_{RGB}$, mask loss $L_{mask}$, and optical flow loss $L_{flow}$. Note that, since the camera is assumed to be fixed, the estimated optical flow will only include scene motion, playing a direct supervisory role. 
However, these losses only add supervision from a single view, making it struggle to ensure the plausibility of other views. We observe that every frame of the 4D NeRF can be considered as a 3D NeRF, 
so to ensure spatial consistency, we utilize the rendering images from the static stage following DreamBooth~\cite{Ruiz2022DreamBoothFT} to fine-tune the diffusion model and introduce bootstrapped score distillation (BSD) loss~\cite{sun2023dreamcraft3d}:
\begin{equation}
\begin{split}
    \nabla_{\theta}L_{BSD}(f(\theta))&=\mathbb{E}_{t,c,\epsilon}[\omega(t)(\epsilon_\text{DreamBooth}(X_{t};y,c,t,r_{{t}^\prime}(X)) \\
    & -\epsilon_\text{lora}(X_{t};y,c,t,X))\frac{\partial X}{\partial \theta}],
\end{split}
\end{equation}
where $\omega(t)$ is the weighting function that depends on the timestep $t$, $r_{{t}^\prime}(X)$ is the augmented renderings used to train the DreamBooth model and $\epsilon_\text{lora}$ estimates the score of the rendered images using a LoRA (Low-rank adaptation)~\cite{hu2021lora} model.
We find that Stable Zero-1-to-3 struggles to maintain multi-view consistency and texture quality in dynamic generation. Therefore, we do not adopt it here.
Finally, in order to eliminate the possible jitters of motion in space and time, we introduce a total variation loss $L_{TV}$ in the image domain, which constrains the rendered 2D displacement map to be similar between adjacent pixels and adjacent times. The overall optimization loss $L_{dynamic}$ in the dynamic stage is:
\begin{align}
    L_{dynamic} &= \lambda_{T} L_{T} + \lambda_{BSD} L_{BSD} + \lambda_{TV} L_{TV} \\ 
    &+ \lambda_{RGB} L_{RGB} + \lambda_{mask} L_{mask} + \lambda_{flow} L_{flow} ,
\end{align}
where $\lambda_{*}$ are the weights for corresponding losses. For the specific definition of all loss terms and the default values of the weights, please refer to the supplementary document.

\subsection{Prior Switching Training Strategy}
\label{sec:training}
In the static stage, we adopt an optimization strategy similar to Magic123~\cite{qian2023magic123}. For the dynamic generation, we have introduced several supervisory losses which can be roughly divided into 3 categories: direct supervision from the reference video, i.e. $L_{RGB}$, $L_{mask}$ and $L_{flow}$ at a fixed initial view, distillation supervision from diffusion models, i.e. $L_{T}$ and $L_{BSD}$, and the regularizer $L_{TV}$. In our experiments, we have found that although direct supervision from the generated video can help stabilize motion generation, the limited dynamic information contained in the generated video gradually limits motion generation, while priors from diffusion models can help enrich motion generation. Therefore, we propose a prior switching training strategy that transitions between the reference video prior and the diffusion model prior. Specifically, at the beginning of the optimization, those losses from the reference video will be given greater loss weights, allowing them to play a main guiding role. Then, as the training step increases, we will gradually reduce the weights of these losses, so that the SDS losses from the diffusion model will play a major supervisory role, promoting the amplitude of the generated motion. It should be noted that since we aim for text-to-4D generation, we do not require the final result to be completely consistent with the reference video, and there may be deviations between the final result and the reference video. In addition, we have empirically found that there are certain difficulties in co-optimizing both video SDS $L_{T}$ and customized SDS loss $L_{BSD}$. So we choose to set an SDS probability and choose which SDS loss to use in each iteration based on the SDS probability. This is also a prior switching strategy between two diffusion priors. Through this training strategy, the final dynamic generation results will not fall into a completely disordered state, resulting in reasonable and sufficient motion.

\section{Experiments and Evaluations}
\label{sec:exp}

\subsection{Evaluation Metrics}
It is not easy to quantitatively evaluate the text-based generated results due to the lack of ground truth. Following \cite{bahmani20234d}, we use two metrics, CLIP score~\cite{radford2021learning} and user survey, for quantitative comparison. CLIP score measures the alignment between the generated results and the input text. Specifically, it extracts the embedding vectors of the text and the image, respectively, and then calculates the cosine similarity between the two vectors. We sample multiple camera trajectories for each method to generate results over time and under changing viewpoints. Then we take the average CLIP score of all frames as the final score. Another metric is user study. We have invited 25 users to evaluate our method and compare to other methods, and the evaluation procedure is the same as \cite{bahmani20234d}. The evaluation criteria are appearance quality (AQ), 3D structure quality (SQ), motion quality (MQ), text alignment (TA), and overall preference (Overall). The reported numbers are the percentages of users who voted for the corresponding method over ours in head-to-head comparisons. 
Please refer to the supplementary document for more details.

\subsection{Results and Comparisons}
Our method is proposed for the text-to-4D generation, but it can also be applied to the monocular-video-to-4D generation task because we introduce the pre-generated reference video to provide direct priors in the generation. The required text prompt can be provided by the user or obtained through some text generation methods~\cite{li2022blip}. Please refer to the supplementary document for more implementation details. Therefore, we compare our method not only with the text-to-4D generation methods, MAV3D~\cite{singer2023text} and recent 4d-fy~\cite{bahmani20234d} but also with the video-to-4D generation method, Consistent4D~\cite{jiang2023consistent4d} and 4DGen~\cite{yin20234dgen}.

\begin{figure}[!t]
    \centering
    \includegraphics[width=0.97\linewidth]{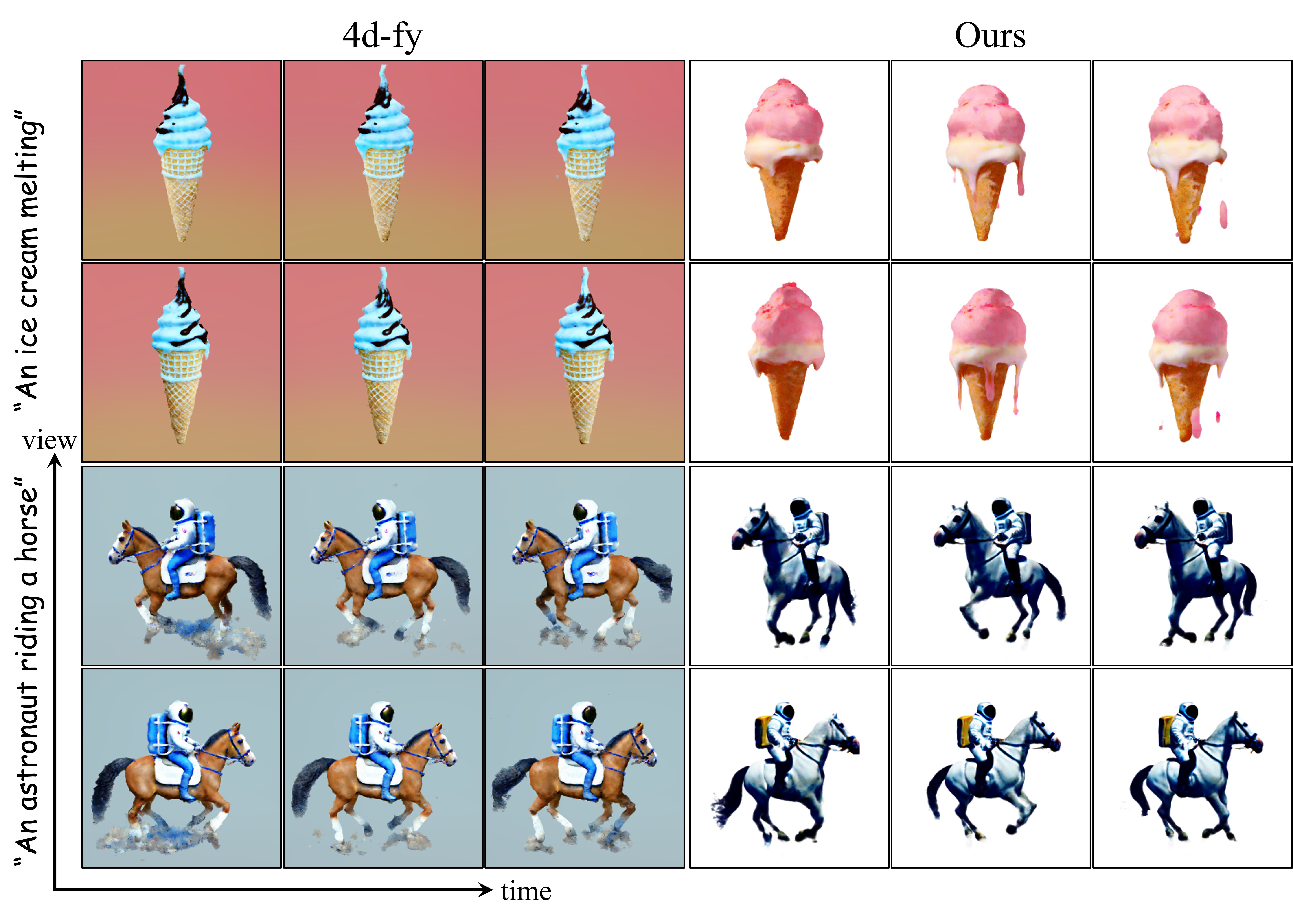}
    \vspace{-2mm}
    \caption{More comparisons of the text-to-4D generation with 4d-fy~\cite{bahmani20234d}. These results demonstrate that our method can generate more pronounced motions, such as the motions of horse legs and outperforms 4d-fy in terms of dynamic generation.}
    \vspace{-3mm}
    \label{fig:compare_4dfy}
\end{figure}

\begin{figure*}[!t]
	\centering
    \includegraphics[width=0.95\linewidth]{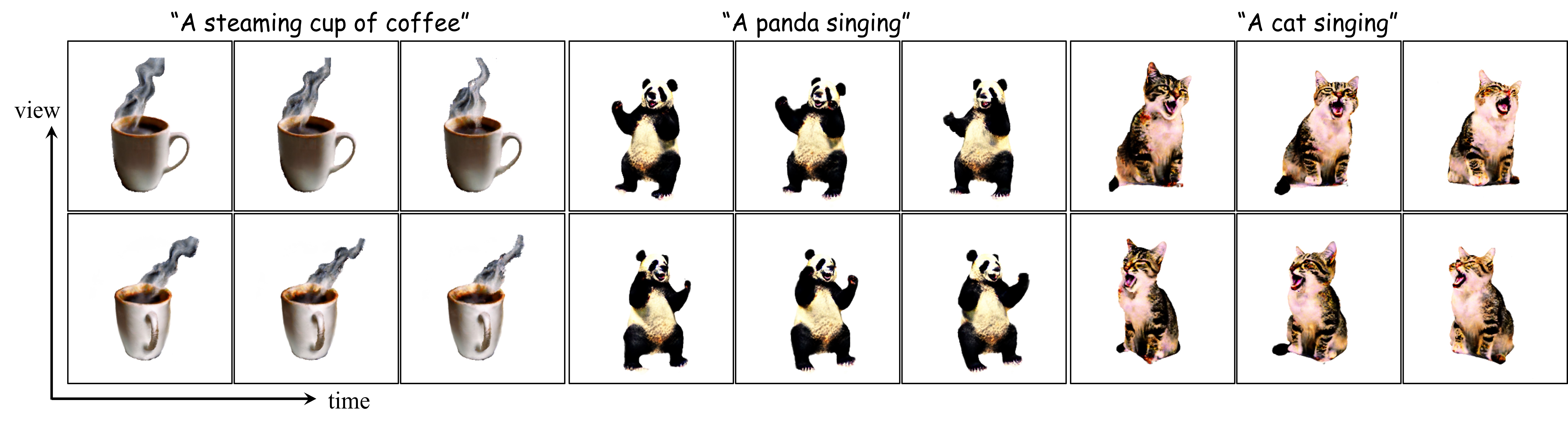} \\
    \includegraphics[width=0.95\linewidth]{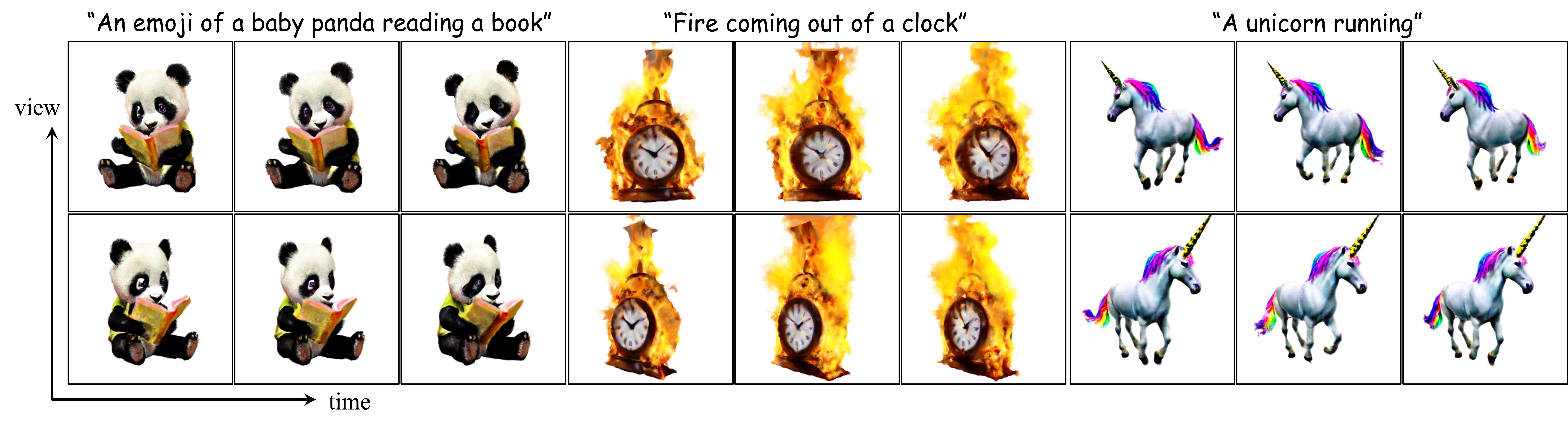}
	\vspace{-2mm}
	\caption{More text-to-4D generation results.}
	\vspace{-2mm}
	\label{fig:more_text_gen}
\end{figure*}

\textbf{Text-to-4D.}
We compare our method with MAV3D~\cite{singer2023text} and 4d-fy~\cite{bahmani20234d} in text-to-4D generation and the qualitative comparisons are shown in Fig.~\ref{fig:text_compare}. It can be seen from the visual comparison that our method has significant advantages over MAV3D, and the generation results are more realistic in both geometry and texture. For example, the panda face and the rocket have more details, and the emitted smoke looks more realistic. 
Compared to the cartoon style of 4d-fy, our results are more realistic, thanks to the introduction of the direct prior from the pre-generated reference video. Meanwhile, our method generates clearer dynamic effects without blurring. Furthermore, in the case of the panda (the first two rows), 4d-fy produces an unreasonable result, with a third leg growing behind the panda's back.
This might be caused by the dynamic representation used by 4d-fy, which is only an additional spatiotemporal encoding and is prone to producing substantial and potentially unreasonable geometric changes to the static model. Our representation includes a deformation network, which can ensure that the generated results are meaningful continuous dynamic effects.
We present more comparison results with 4d-fy in Fig.~\ref{fig:compare_4dfy}. These results illustrate that the dynamic amplitude of our method is higher than 4d-fy. The supplementary video provides a better viewing experience.
We also present quantitative results in Table~\ref{tab:text}. The user study proves that users tend to prefer our method. 
More text-to-4D generation results are shown in Fig.~\ref{fig:more_text_gen}.

\begin{figure*}[htb]
	\centering
	\includegraphics[width=0.95\linewidth]{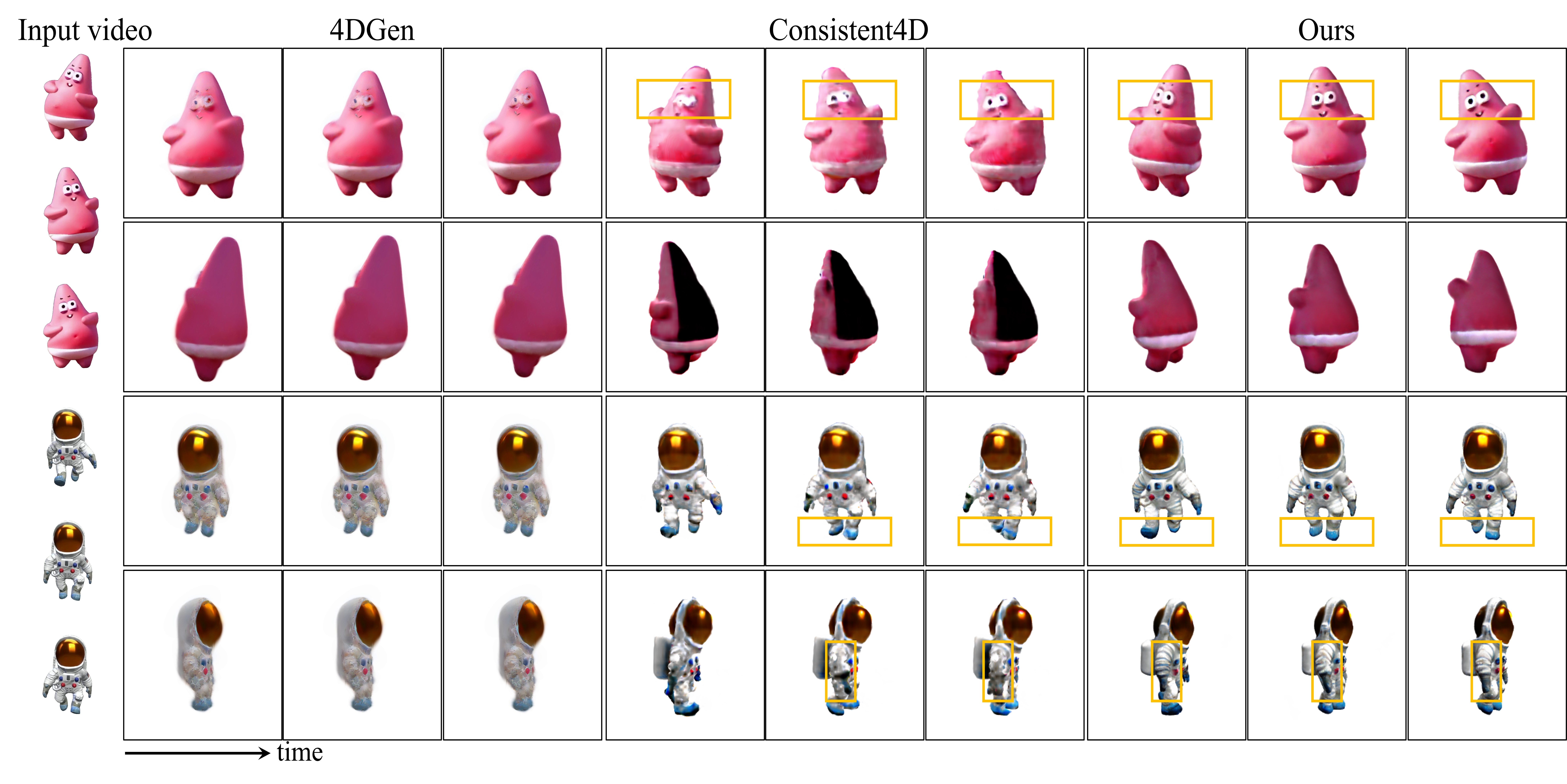} \\
	\vspace{-2mm}
	\caption{Comparisons of the video-to-4D generation with Consistent4D and 4DGen. We show the corresponding input frames in the left column. Our method maintains the texture well under the novel view and produces more vivid results.}
	\vspace{-2mm}
	\label{fig:video_compare}
\end{figure*}

\begin{figure*}[!t]
	\centering
    \includegraphics[width=0.95\linewidth]{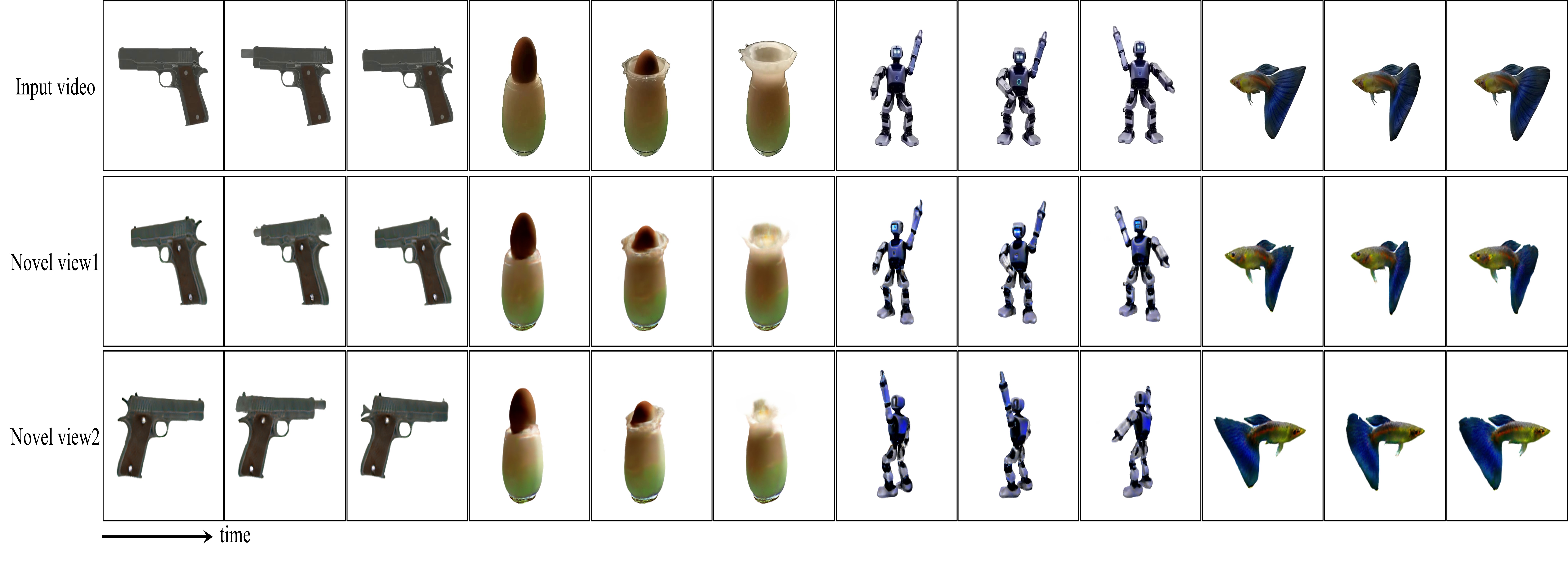}
	\vspace{-4mm}
	\caption{More video-to-4D generation results.}
	\vspace{-2mm}
	\label{fig:more_video_gen}
\end{figure*}

\textbf{Video-to-4D.}
We visualize the 4D generation results from monocular videos in Fig.~\ref{fig:video_compare} and also compare with the results of Consistent4D~\cite{jiang2023consistent4d} and 4DGen~\cite{yin20234dgen}. 
The first group of comparisons shows that our method maintains the texture well under the novel view where the white diaper of Patrick Star disappears from the results of Consistent4D. And the astronaut example shows that our method produces more vivid results and has richer details, while 4DGen fails to accurately reconstruct the appearance and motions. We show more video-to-4D generation results in Fig.~\ref{fig:more_video_gen}.

\begin{table}[!t]
    \caption{The quantitative comparisons with MAV3D, 4d-fy, and variations of our method. The methods are evaluated in terms of CLIP Score (CLIP) and those metrics of user study where each alternative method is compared with ours.}
    \vspace{-6mm}
    \label{tab:text}
    \begin{center}
    \resizebox{\columnwidth}{!}{
    \begin{tabular}{lc|cccc|c}
        \toprule
         &  & \multicolumn{5}{c}{\textit{User Study}}\\
        \textit{Method} & \textit{CLIP} & AQ & SQ & MQ & TA & Overall \\\midrule
        MAV3D & 29.4 & 36\% & 40\% & 48\% & 36\% & 36\% \\
        4D-fy & 30.2 & 44\% & 44\% & 44\% & 48\% & 44\% \\
        Ours  & 31.7 & \multicolumn{4}{c|}{---} & ---\\
        \midrule
        \textit{Ablation Study} & \multicolumn{6}{c}{}  \\\midrule
        w/o topology network & 30.0 & 32\% & 36\% & 24\% & 28\% & 28\% \\
        w/o direct prior     & 29.7 & 20\% & 8\% & 12\% & 12\% & 8\% \\
        w/o prior-switching  & 29.1 & 36\% & 36\% & 48\% & 36\% & 48\% \\
        w/o 3D SDS           & 29.1 & 8\% & 0\% & 8\% & 8\% & 0\% \\%\midrule
        w/o video SDS        & 29.4 & 36\% & 28\% & 28\% & 24\% & 28\% \\
        Ours                 & 31.0 & \multicolumn{4}{c|}{---} & ---\\
        \bottomrule
    \end{tabular}}
    \end{center}
    \vspace{-4mm}
\end{table}

\subsection{Ablation Study}
\begin{figure*}[!t]
    \centering
    \includegraphics[width=0.97\linewidth]{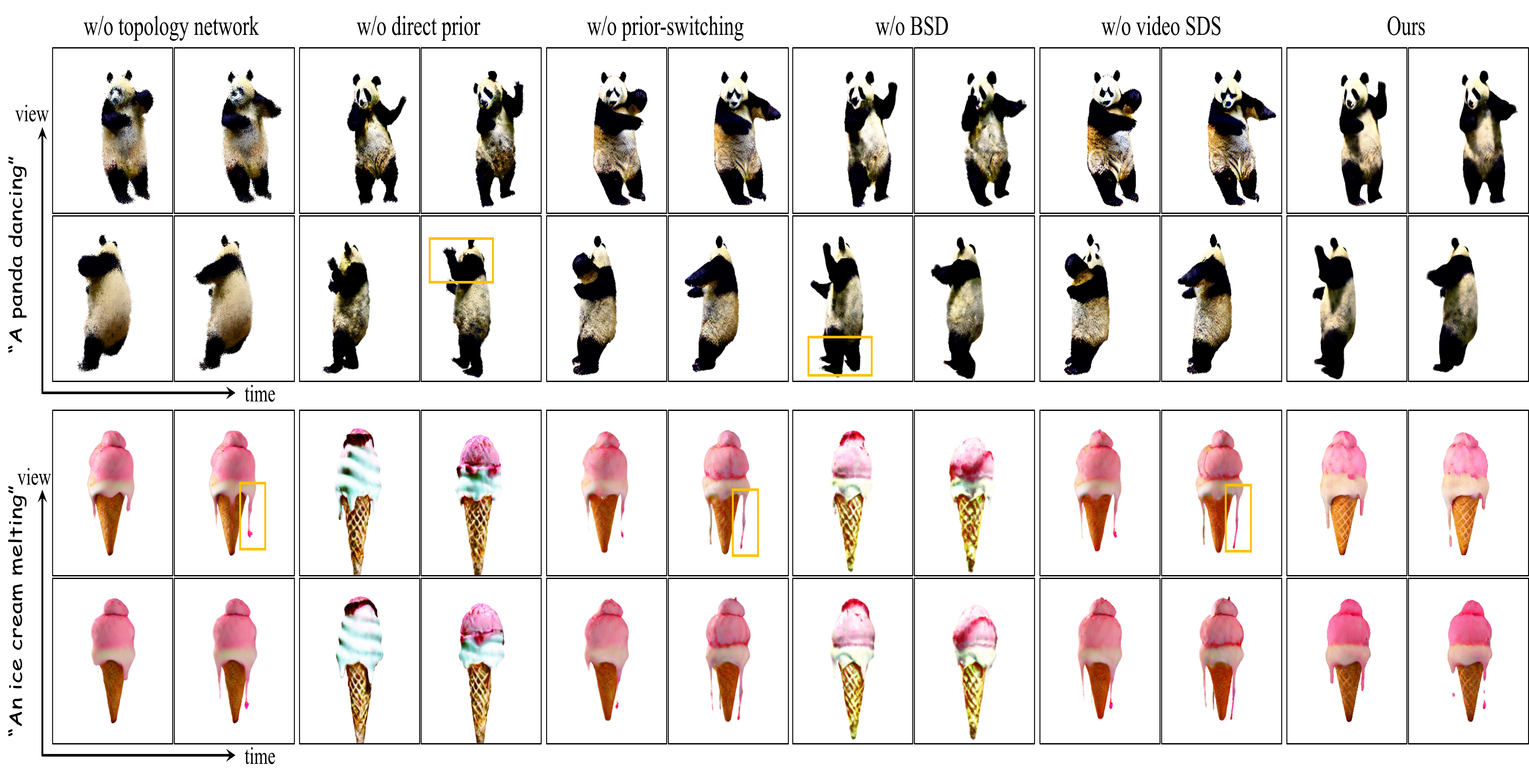}
    \vspace{-3mm}
    \caption{Ablation study. Without the topology network, the 4D representation loses some modeling ability, leading to noise in the results. Lack of direct prior can lead to uncontrollable optimization and blurry outcomes. Lack of the prior-switching strategy keeps direct prior dominant, making the results affected by the quality of the generated reference video and unsatisfactory. Lack of BSD loss will affect 3D consistency. For example, the panda has three legs in some views. The lack of video SDS is similar to the lack of prior-switching strategy as the generated reference video guides the optimization which limits the generation and results in poor results. Our method can achieve optimal results in both appearance and dynamics. 
    In the case of the melting ice cream, the full version of our method generates the melting ice cream with a clearer appearance and a more realistic ice cream dropping effect compared to alternative methods.}
    \vspace{-1mm}
    \label{fig:ablation}
\end{figure*}

To verify the effectiveness of the dynamic representation, 4D generation pipeline, and training strategy used in our method, we perform multiple ablation experiments. The qualitative and quantitative results are shown in Fig.~\ref{fig:ablation} and Table~\ref{tab:text}, respectively.

\begin{figure}[!t]
    \centering
    \includegraphics[width=0.97\linewidth]{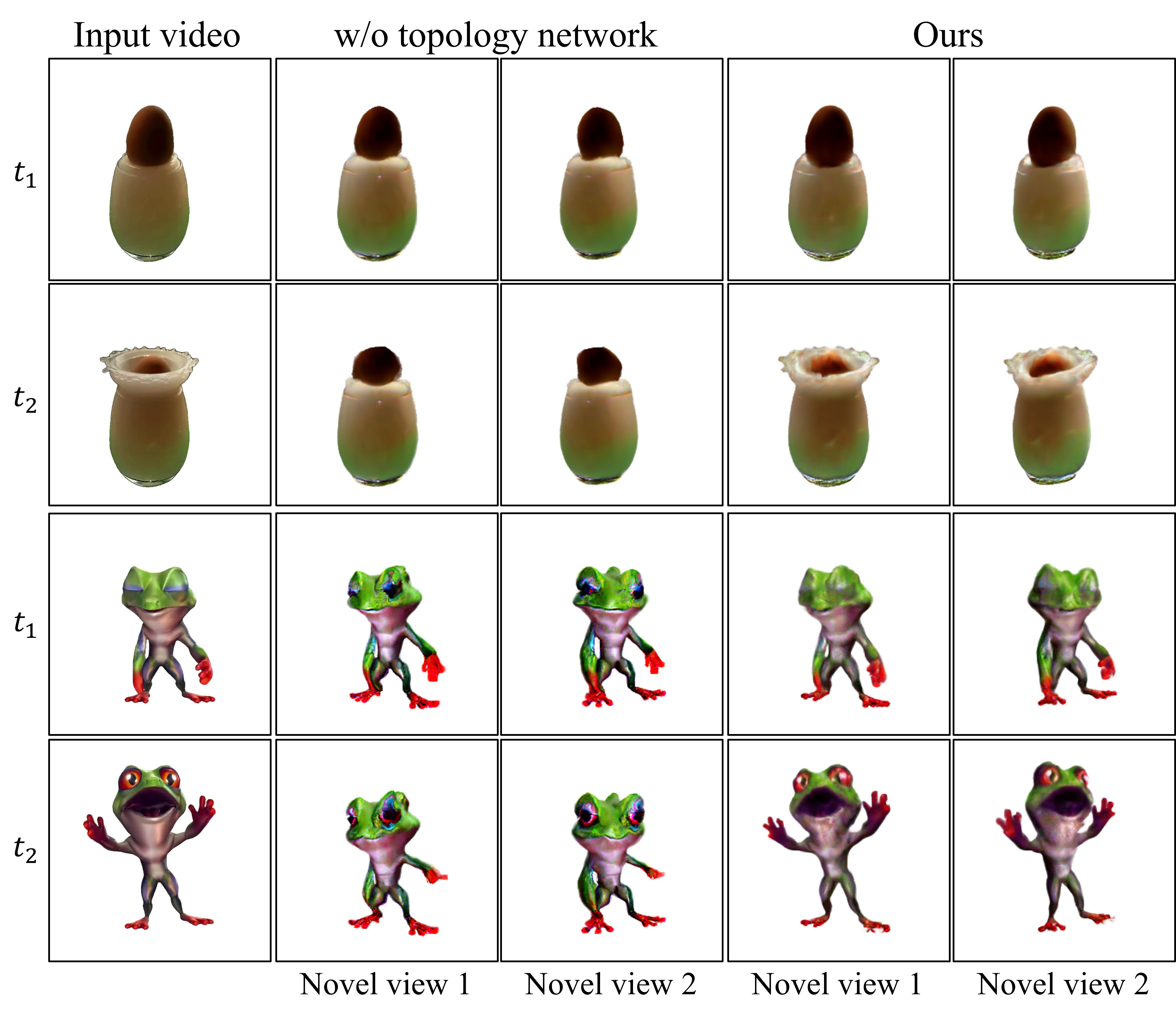}
    \vspace{-3mm}
    \caption{Ablation study for the topology network. We choose video-to-4D cases, which are more controllable, to illustrate the function of the topology network. As can be seen, without the topology network, we cannot achieve dynamic generation of an egg falling into a cup filled with liquid and a frog opening its eyes and mouth, which involve substantial topology changes.}
    \vspace{-3mm}
    \label{fig:ablation_topo}
\end{figure}

\textbf{w/o topology network.}
Our dynamic representation not only uses the deformation network but also introduces the topology network to ensure the continuity and diversity of motion. The introduction of the deformation network is a natural idea, so we conduct an ablation experiment on the topology network to verify whether it will affect the dynamic generation effect. We show two video-to-4D examples in Fig.~\ref{fig:ablation_topo} to illustrate the function of the topology network in some specific dynamics. In the examples, an egg is falling into a cup filled with liquid and a frog is opening its eyes and mouth. As can be seen, without the topology network, these dynamics cannot be achieved. Fig.~\ref{fig:ablation} further shows text-to-4D examples where the generation effect with only the deformation network is unsatisfactory.

\textbf{w/o direct prior.}
The core idea of our method is to use the direct prior of the pre-generated reference video to guide tex-to-4D generation. Because we focus on the dynamic generation, we remove the direct losses from the reference video during the dynamic generation stage. Although the reference video is not a hard constraint, the lack of direct prior guidance in the early stages of optimization will lead dynamic generation towards an unreasonable outcome, resulting in worse results. Note the area marked by the orange box in the first case of Fig.~\ref{fig:ablation}.

\textbf{w/o prior-switching.}
The reference video may not contain sufficient dynamic information. So to exploit the generation ability of SDS distillation, we design a prior-switching training strategy to gradually reduce the weights of the direct losses. If this training strategy is not adopted, there will be conflicts between the direct prior and the generative prior from SDS in the later stage of training. The results shown in Fig.~\ref{fig:ablation} illustrate this point.

\textbf{w/o BSD.}
The customized loss function, i.e. BSD loss, helps maintain 3D consistency and texture quality. Removing BSD loss will result in a decrease in generation quality and inconsistent multi-views. For example, in the `w/o BSD' column of Fig.~\ref{fig:ablation}, the panda has three legs in some views and the texture of the generated ice cream is very poor.

\textbf{w/o video SDS.}
The video SDS loss function extracts dynamic information from the 2D video diffusion model to ensure dynamic continuity. Without using this loss, reasonable dynamic generation results cannot be obtained.

\begin{figure}[!t]
    \centering
    \includegraphics[width=0.97\linewidth]{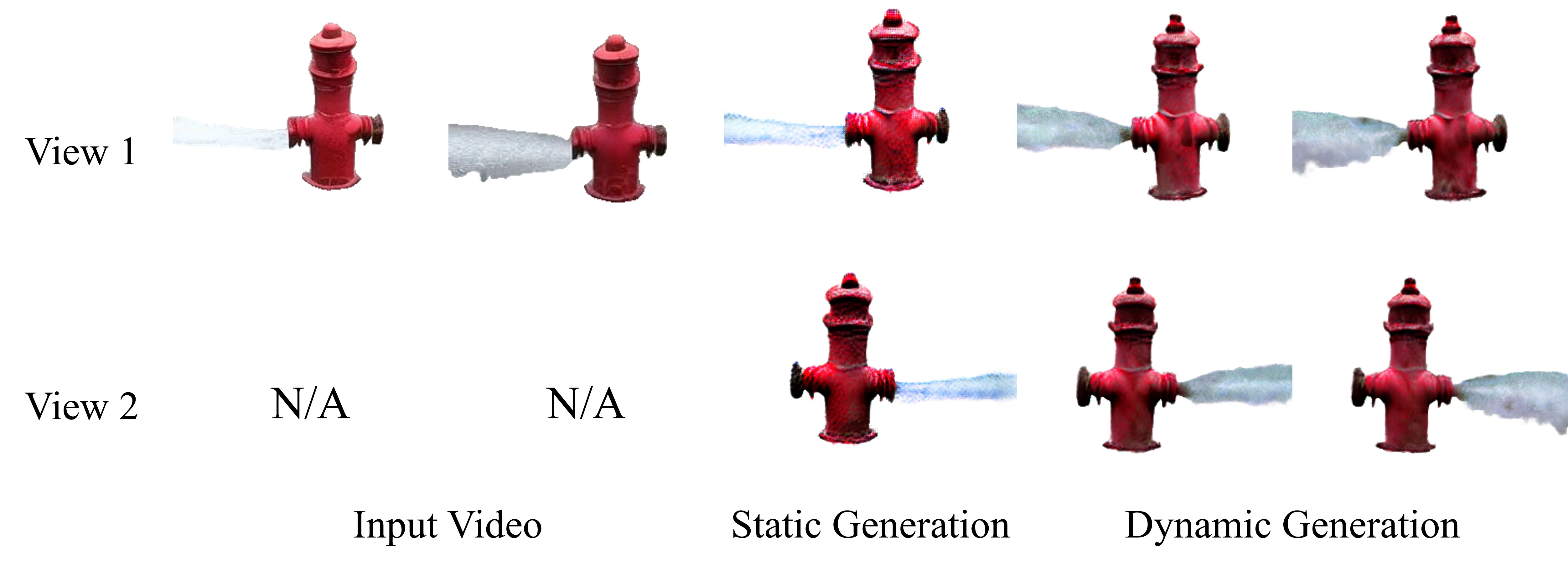}
    \vspace{-3mm}
    \caption{A failure case. We show an example to illustrate how our method is affected by the generated reference video and static 3D generation result. In this case, the static generation affected the fire hydrant with a reduced quality appearance, and the generated video affected the water which looks less realistic.}
    \vspace{-3mm}
    \label{fig:failure_case}
\end{figure}

\section{Discussions and Conclusions}
\label{sec:conclusion}

\textbf{Limitations.} 
Text-to-4D generation is a challenging task, so our method, as an initial attempt, still has some limitations. First, we exploit the image-to-3D generation model to generate the static shape as the basis for dynamic generation. However, texture inaccuracy and Janus problems exist in some of the reconstructed 3D models. Although our 4D generation results ultimately are not bound to the geometry and texture of the static model, its use as an initial value for optimization will still affect the final generation quality. Likewise, our method utilizes the direct prior from the generated reference video to ensure generation quality and motion amplitude and the current text-to-video generative model still faces issues such as geometric mutations over time, which may influence the final effects of our method. An example failure case is shown in Fig.~\ref{fig:failure_case}. Given the rapid development in image-to-3D and text-to-video fields, our method will benefit from the development of these methods to achieve higher-quality results. Moreover, some difficult-to-reconstruct objects, such as transparent objects, cannot appear in dynamic generation.
Second, our method adopts the SDS optimization for generation, which is slower than feed-forward generation. And we use the NeRF representation, which makes the entire optimization generation process last several hours. Since it is difficult to collect a corresponding paired dataset to train a feed-forward generation model, we will explore replacing the NeRF with the efficient 3D Gaussian Splatting representation~\cite{kerbl20233d} to improve the generation speed.

\textbf{Ethical issues.}
As a generation method, our method may be abused, generating misleading or false information. We will carefully control open access rights and put warning labels on the generation results.

In this paper, we propose a novel text-to-4D generation method that fully utilizes the priors from the text-to-video diffusion model and the pre-generated reference video. First, we introduce a hybrid dynamic representation of a deformation network and a topology network in 4D generation, modeling dynamic continuity and topological changes. Then, we divide the generation process into two stages (static and dynamic) and propose to adopt the pre-generated reference video to provide direct priors, combined with customized and video-based SDS losses, to achieve high-quality 4D generation. For this purpose, we also design a prior-switching training strategy to balance the direct prior of the reference video and the generative prior of the diffusion model. In a user study and quality comparisons, our method has proven to outperform existing methods.

\ifCLASSOPTIONcaptionsoff
  \newpage
\fi

{\small
	\bibliographystyle{IEEEtran}
	\bibliography{bibliography}

\begin{thebibliography}{100}
\providecommand{\url}[1]{#1}
\csname url@samestyle\endcsname
\providecommand{\newblock}{\relax}
\providecommand{\bibinfo}[2]{#2}
\providecommand{\BIBentrySTDinterwordspacing}{\spaceskip=0pt\relax}
\providecommand{\BIBentryALTinterwordstretchfactor}{4}
\providecommand{\BIBentryALTinterwordspacing}{\spaceskip=\fontdimen2\font plus
\BIBentryALTinterwordstretchfactor\fontdimen3\font minus \fontdimen4\font\relax}
\providecommand{\BIBforeignlanguage}[2]{{%
\expandafter\ifx\csname l@#1\endcsname\relax
\typeout{** WARNING: IEEEtran.bst: No hyphenation pattern has been}%
\typeout{** loaded for the language `#1'. Using the pattern for}%
\typeout{** the default language instead.}%
\else
\language=\csname l@#1\endcsname
\fi
#2}}
\providecommand{\BIBdecl}{\relax}
\BIBdecl

\bibitem{3dgan}
J.~Wu, C.~Zhang, T.~Xue, B.~Freeman, and J.~Tenenbaum, ``Learning a probabilistic latent space of object shapes via 3d generative-adversarial modeling,'' in \emph{Advances in Neural Information Processing Systems}, 2016, pp. 82--90.

\bibitem{spgn}
H.~Fan, H.~Su, and L.~J. Guibas, ``A point set generation network for 3d object reconstruction from a single image,'' in \emph{CVPR}, 2017, pp. 2463--2471.

\bibitem{SDM-NET}
L.~Gao, J.~Yang, T.~Wu, Y.~Yuan, H.~Fu, Y.~Lai, and H.~Zhang, ``{SDM-NET:} deep generative network for structured deformable mesh,'' \emph{{ACM} Trans. Graph.}, vol.~38, no.~6, pp. 243:1--243:15, 2019.

\bibitem{yang2022dsg}
J.~Yang, K.~Mo, Y.-K. Lai, L.~J. Guibas, and L.~Gao, ``Dsg-net: Learning disentangled structure and geometry for 3d shape generation,'' \emph{ACM Transactions on Graphics (TOG)}, vol.~42, no.~1, pp. 1--17, 2022.

\bibitem{deitke2023objaverse}
M.~Deitke, D.~Schwenk, J.~Salvador, L.~Weihs, O.~Michel, E.~VanderBilt, L.~Schmidt, K.~Ehsani, A.~Kembhavi, and A.~Farhadi, ``Objaverse: A universe of annotated 3d objects,'' in \emph{Proceedings of the IEEE/CVF Conference on Computer Vision and Pattern Recognition}, 2023, pp. 13\,142--13\,153.

\bibitem{deitke2023objaversexl}
M.~Deitke, R.~Liu, M.~Wallingford, H.~Ngo, O.~Michel, A.~Kusupati, A.~Fan, C.~Laforte, V.~Voleti, S.~Y. Gadre \emph{et~al.}, ``Objaverse-xl: A universe of 10m+ 3d objects,'' \emph{arXiv preprint arXiv:2307.05663}, 2023.

\bibitem{mildenhall2020nerf}
B.~Mildenhall, P.~P. Srinivasan, M.~Tancik, J.~T. Barron, R.~Ramamoorthi, and R.~Ng, ``{NeRF}: Representing scenes as neural radiance fields for view synthesis,'' in \emph{European Conference on Computer Vision}.\hskip 1em plus 0.5em minus 0.4em\relax Springer, 2020, pp. 405--421.

\bibitem{eg3d}
\BIBentryALTinterwordspacing
E.~R. Chan, C.~Z. Lin, M.~A. Chan, K.~Nagano, B.~Pan, S.~D. Mello, O.~Gallo, L.~J. Guibas, J.~Tremblay, S.~Khamis, T.~Karras, and G.~Wetzstein, ``Efficient geometry-aware 3d generative adversarial networks,'' in \emph{{IEEE/CVF} Conference on Computer Vision and Pattern Recognition, {CVPR} 2022, New Orleans, LA, USA, June 18-24, 2022}.\hskip 1em plus 0.5em minus 0.4em\relax {IEEE}, 2022, pp. 16\,102--16\,112. [Online]. Available: \url{https://doi.org/10.1109/CVPR52688.2022.01565}
\BIBentrySTDinterwordspacing

\bibitem{liu2020neural}
L.~Liu, J.~Gu, K.~Zaw~Lin, T.-S. Chua, and C.~Theobalt, ``Neural sparse voxel fields,'' \emph{Advances in Neural Information Processing Systems}, vol.~33, 2020.

\bibitem{mueller2022instant}
\BIBentryALTinterwordspacing
T.~M\"uller, A.~Evans, C.~Schied, and A.~Keller, ``Instant neural graphics primitives with a multiresolution hash encoding,'' \emph{ACM Trans. Graph.}, vol.~41, no.~4, pp. 102:1--102:15, Jul. 2022. [Online]. Available: \url{https://doi.org/10.1145/3528223.3530127}
\BIBentrySTDinterwordspacing

\bibitem{karras2019style}
T.~Karras, S.~Laine, and T.~Aila, ``A style-based generator architecture for generative adversarial networks,'' in \emph{Proceedings of the IEEE/CVF conference on computer vision and pattern recognition}, 2019, pp. 4401--4410.

\bibitem{rombach2022high}
R.~Rombach, A.~Blattmann, D.~Lorenz, P.~Esser, and B.~Ommer, ``High-resolution image synthesis with latent diffusion models,'' in \emph{Proceedings of the IEEE/CVF conference on computer vision and pattern recognition}, 2022, pp. 10\,684--10\,695.

\bibitem{gao2022get3d}
J.~Gao, T.~Shen, Z.~Wang, W.~Chen, K.~Yin, D.~Li, O.~Litany, Z.~Gojcic, and S.~Fidler, ``Get3d: A generative model of high quality 3d textured shapes learned from images,'' \emph{Advances In Neural Information Processing Systems}, vol.~35, pp. 31\,841--31\,854, 2022.

\bibitem{wang2023rodin}
T.~Wang, B.~Zhang, T.~Zhang, S.~Gu, J.~Bao, T.~Baltrusaitis, J.~Shen, D.~Chen, F.~Wen, Q.~Chen \emph{et~al.}, ``Rodin: A generative model for sculpting 3d digital avatars using diffusion,'' in \emph{Proceedings of the IEEE/CVF Conference on Computer Vision and Pattern Recognition}, 2023, pp. 4563--4573.

\bibitem{chen2023single}
H.~Chen, J.~Gu, A.~Chen, W.~Tian, Z.~Tu, L.~Liu, and H.~Su, ``Single-stage diffusion nerf: A unified approach to 3d generation and reconstruction,'' \emph{arXiv preprint arXiv:2304.06714}, 2023.

\bibitem{karnewar2023holodiffusion}
A.~Karnewar, A.~Vedaldi, D.~Novotny, and N.~J. Mitra, ``Holodiffusion: Training a 3d diffusion model using 2d images,'' in \emph{Proceedings of the IEEE/CVF Conference on Computer Vision and Pattern Recognition}, 2023, pp. 18\,423--18\,433.

\bibitem{karnewar2023holofusion}
A.~Karnewar, N.~J. Mitra, A.~Vedaldi, and D.~Novotny, ``Holofusion: Towards photo-realistic 3d generative modeling,'' in \emph{Proceedings of the IEEE/CVF International Conference on Computer Vision}, 2023, pp. 22\,976--22\,985.

\bibitem{schuhmann2022laion}
C.~Schuhmann, R.~Beaumont, R.~Vencu, C.~Gordon, R.~Wightman, M.~Cherti, T.~Coombes, A.~Katta, C.~Mullis, M.~Wortsman \emph{et~al.}, ``Laion-5b: An open large-scale dataset for training next generation image-text models,'' \emph{Advances in Neural Information Processing Systems}, vol.~35, pp. 25\,278--25\,294, 2022.

\bibitem{poole2022dreamfusion}
B.~Poole, A.~Jain, J.~T. Barron, and B.~Mildenhall, ``Dreamfusion: Text-to-3d using 2d diffusion,'' \emph{arXiv preprint arXiv:2209.14988}, 2022.

\bibitem{liu2023zero}
R.~Liu, R.~Wu, B.~Van~Hoorick, P.~Tokmakov, S.~Zakharov, and C.~Vondrick, ``Zero-1-to-3: Zero-shot one image to 3d object,'' in \emph{Proceedings of the IEEE/CVF International Conference on Computer Vision}, 2023, pp. 9298--9309.

\bibitem{shi2023mvdream}
Y.~Shi, P.~Wang, J.~Ye, M.~Long, K.~Li, and X.~Yang, ``Mvdream: Multi-view diffusion for 3d generation,'' \emph{arXiv preprint arXiv:2308.16512}, 2023.

\bibitem{wang2023prolificdreamer}
Z.~Wang, C.~Lu, Y.~Wang, F.~Bao, C.~Li, H.~Su, and J.~Zhu, ``Prolificdreamer: High-fidelity and diverse text-to-3d generation with variational score distillation,'' \emph{arXiv preprint arXiv:2305.16213}, 2023.

\bibitem{yu2023text}
X.~Yu, Y.-C. Guo, Y.~Li, D.~Liang, S.-H. Zhang, and X.~Qi, ``Text-to-3d with classifier score distillation,'' \emph{arXiv preprint arXiv:2310.19415}, 2023.

\bibitem{chen2023fantasia3d}
R.~Chen, Y.~Chen, N.~Jiao, and K.~Jia, ``Fantasia3d: Disentangling geometry and appearance for high-quality text-to-3d content creation,'' \emph{arXiv preprint arXiv:2303.13873}, 2023.

\bibitem{singer2023text}
U.~Singer, S.~Sheynin, A.~Polyak, O.~Ashual, I.~Makarov, F.~Kokkinos, N.~Goyal, A.~Vedaldi, D.~Parikh, J.~Johnson \emph{et~al.}, ``Text-to-4d dynamic scene generation,'' \emph{arXiv preprint arXiv:2301.11280}, 2023.

\bibitem{cao2023hexplane}
A.~Cao and J.~Johnson, ``Hexplane: A fast representation for dynamic scenes,'' in \emph{Proceedings of the IEEE/CVF Conference on Computer Vision and Pattern Recognition}, 2023, pp. 130--141.

\bibitem{wang2021neus}
P.~Wang, L.~Liu, Y.~Liu, C.~Theobalt, T.~Komura, and W.~Wang, ``{NeuS}: Learning neural implicit surfaces by volume rendering for multi-view reconstruction,'' in \emph{Advances in Neural Information Processing Systems}, vol.~34, 2021.

\bibitem{peng2021neural}
S.~Peng, Y.~Zhang, Y.~Xu, Q.~Wang, Q.~Shuai, H.~Bao, and X.~Zhou, ``Neural body: Implicit neural representations with structured latent codes for novel view synthesis of dynamic humans,'' in \emph{Proceedings of the IEEE/CVF Conference on Computer Vision and Pattern Recognition}, 2021, pp. 9054--9063.

\bibitem{zhang2021editable}
J.~Zhang, X.~Liu, X.~Ye, F.~Zhao, Y.~Zhang, M.~Wu, Y.~Zhang, L.~Xu, and J.~Yu, ``Editable free-viewpoint video using a layered neural representation,'' \emph{ACM Transactions on Graphics (TOG)}, vol.~40, no.~4, pp. 1--18, 2021.

\bibitem{liu2021neural}
L.~Liu, M.~Habermann, V.~Rudnev, K.~Sarkar, J.~Gu, and C.~Theobalt, ``Neural actor: Neural free-view synthesis of human actors with pose control,'' \emph{ACM transactions on graphics (TOG)}, vol.~40, no.~6, pp. 1--16, 2021.

\bibitem{weng2022humannerf}
C.-Y. Weng, B.~Curless, P.~P. Srinivasan, J.~T. Barron, and I.~Kemelmacher-Shlizerman, ``Humannerf: Free-viewpoint rendering of moving people from monocular video,'' in \emph{Proceedings of the IEEE/CVF conference on computer vision and pattern Recognition}, 2022, pp. 16\,210--16\,220.

\bibitem{jiang2022neuman}
W.~Jiang, K.~M. Yi, G.~Samei, O.~Tuzel, and A.~Ranjan, ``Neuman: Neural human radiance field from a single video,'' in \emph{European Conference on Computer Vision}.\hskip 1em plus 0.5em minus 0.4em\relax Springer, 2022, pp. 402--418.

\bibitem{liu2023hosnerf}
J.-W. Liu, Y.-P. Cao, T.~Yang, Z.~Xu, J.~Keppo, Y.~Shan, X.~Qie, and M.~Z. Shou, ``Hosnerf: Dynamic human-object-scene neural radiance fields from a single video,'' in \emph{Proceedings of the IEEE/CVF International Conference on Computer Vision}, 2023, pp. 18\,483--18\,494.

\bibitem{barron2021mip}
J.~T. Barron, B.~Mildenhall, M.~Tancik, P.~Hedman, R.~Martin-Brualla, and P.~P. Srinivasan, ``Mip-nerf: A multiscale representation for anti-aliasing neural radiance fields,'' in \emph{Proceedings of the IEEE/CVF International Conference on Computer Vision}, 2021, pp. 5855--5864.

\bibitem{hu2023tri}
W.~Hu, Y.~Wang, L.~Ma, B.~Yang, L.~Gao, X.~Liu, and Y.~Ma, ``Tri-miprf: Tri-mip representation for efficient anti-aliasing neural radiance fields,'' in \emph{Proceedings of the IEEE/CVF International Conference on Computer Vision}, 2023, pp. 19\,774--19\,783.

\bibitem{wang2021ibrnet}
Q.~Wang, Z.~Wang, K.~Genova, P.~Srinivasan, H.~Zhou, J.~T. Barron, R.~Martin-Brualla, N.~Snavely, and T.~Funkhouser, ``{IBRNet}: Learning multi-view image-based rendering,'' in \emph{Proceedings of the IEEE/CVF Conference on Computer Vision and Pattern Recognition}, 2021, pp. 4690--4699.

\bibitem{chen2021mvsnerf}
A.~Chen, Z.~Xu, F.~Zhao, X.~Zhang, F.~Xiang, J.~Yu, and H.~Su, ``{MVSNeRF}: Fast generalizable radiance field reconstruction from multi-view stereo,'' in \emph{Proceedings of the IEEE/CVF International Conference on Computer Vision}, 2021, pp. 14\,124--14\,133.

\bibitem{garbin2021fastnerf}
S.~J. Garbin, M.~Kowalski, M.~Johnson, J.~Shotton, and J.~Valentin, ``{FastNeRF}: High-fidelity neural rendering at 200fps,'' in \emph{Proceedings of the IEEE/CVF International Conference on Computer Vision}, 2021, pp. 14\,346--14\,355.

\bibitem{yu2021plenoctrees}
A.~Yu, R.~Li, M.~Tancik, H.~Li, R.~Ng, and A.~Kanazawa, ``Plenoctrees for real-time rendering of neural radiance fields,'' in \emph{Proceedings of the IEEE/CVF International Conference on Computer Vision}, 2021, pp. 5752--5761.

\bibitem{hedman2021baking}
P.~Hedman, P.~P. Srinivasan, B.~Mildenhall, J.~T. Barron, and P.~Debevec, ``Baking neural radiance fields for real-time view synthesis,'' in \emph{Proceedings of the IEEE/CVF International Conference on Computer Vision}, 2021, pp. 5875--5884.

\bibitem{Chen2022tensorf}
A.~Chen, Z.~Xu, A.~Geiger, J.~Yu, and H.~Su, ``Tensorf: Tensorial radiance fields,'' in \emph{European Conference on Computer Vision (ECCV)}, 2022.

\bibitem{kerbl20233d}
B.~Kerbl, G.~Kopanas, T.~Leimk{\"u}hler, and G.~Drettakis, ``3d gaussian splatting for real-time radiance field rendering,'' \emph{ACM Transactions on Graphics}, vol.~42, no.~4, 2023.

\bibitem{liu2021editing}
S.~Liu, X.~Zhang, Z.~Zhang, R.~Zhang, J.-Y. Zhu, and B.~Russell, ``Editing conditional radiance fields,'' in \emph{Proceedings of the IEEE/CVF International Conference on Computer Vision}, 2021, pp. 5773--5783.

\bibitem{yuan2022nerf}
Y.-J. Yuan, Y.-T. Sun, Y.-K. Lai, Y.~Ma, R.~Jia, and L.~Gao, ``Nerf-editing: geometry editing of neural radiance fields,'' in \emph{Proceedings of the IEEE/CVF Conference on Computer Vision and Pattern Recognition}, 2022, pp. 18\,353--18\,364.

\bibitem{huang2022stylizednerf}
Y.-H. Huang, Y.~He, Y.-J. Yuan, Y.-K. Lai, and L.~Gao, ``{StylizedNeRF}: consistent {3D} scene stylization as stylized {NeRF} via {2D-3D} mutual learning,'' in \emph{Proceedings of the IEEE/CVF Conference on Computer Vision and Pattern Recognition}, 2022, pp. 18\,342--18\,352.

\bibitem{neumesh}
{Chong Bao and Bangbang Yang}, Z.~Junyi, B.~Hujun, Z.~Yinda, C.~Zhaopeng, and Z.~Guofeng, ``{NeuMesh}: Learning disentangled neural mesh-based implicit field for geometry and texture editing,'' in \emph{European Conference on Computer Vision (ECCV)}, 2022.

\bibitem{NerfFaceEditing}
K.~Jiang, S.-Y. Chen, F.-L. Liu, H.~Fu, and L.~Gao, ``Nerffaceediting: Disentangled face editing in neural radiance fields,'' in \emph{ACM SIGGRAPH Asia 2022 Conference Proceedings}, ser. SIGGRAPH Asia'22.\hskip 1em plus 0.5em minus 0.4em\relax New York, NY, USA: Association for Computing Machinery, 2022.

\bibitem{dellaert2020neural}
F.~Dellaert and L.~Yen-Chen, ``Neural volume rendering: Nerf and beyond,'' \emph{arXiv preprint arXiv:2101.05204}, 2020.

\bibitem{gao2022nerf}
K.~Gao, Y.~Gao, H.~He, D.~Lu, L.~Xu, and J.~Li, ``Nerf: Neural radiance field in 3d vision, a comprehensive review,'' \emph{arXiv preprint arXiv:2210.00379}, 2022.

\bibitem{tewari2022advances}
A.~Tewari, J.~Thies, B.~Mildenhall, P.~Srinivasan, E.~Tretschk, W.~Yifan, C.~Lassner, V.~Sitzmann, R.~Martin-Brualla, S.~Lombardi \emph{et~al.}, ``Advances in neural rendering,'' in \emph{Computer Graphics Forum}, vol.~41, no.~2.\hskip 1em plus 0.5em minus 0.4em\relax Wiley Online Library, 2022, pp. 703--735.

\bibitem{pumarola2021d}
A.~Pumarola, E.~Corona, G.~Pons-Moll, and F.~Moreno-Noguer, ``{D-NeRF}: Neural radiance fields for dynamic scenes,'' in \emph{Proceedings of the IEEE/CVF Conference on Computer Vision and Pattern Recognition}, 2021, pp. 10\,318--10\,327.

\bibitem{tretschk2021non}
E.~Tretschk, A.~Tewari, V.~Golyanik, M.~Zollhofer, C.~Lassner, and C.~Theobalt, ``Non-rigid neural radiance fields: Reconstruction and novel view synthesis of a dynamic scene from monocular video,'' in \emph{Proceedings of the IEEE/CVF International Conference on Computer Vision}, 2021, pp. 12\,959--12\,970.

\bibitem{li2021nsff}
Z.~Li, S.~Niklaus, N.~Snavely, and O.~Wang, ``Neural scene flow fields for space-time view synthesis of dynamic scenes,'' in \emph{Proceedings of the IEEE/CVF Conference on Computer Vision and Pattern Recognition}, 2021, pp. 6498--6508.

\bibitem{xian2021space}
W.~Xian, J.-B. Huang, J.~Kopf, and C.~Kim, ``Space-time neural irradiance fields for free-viewpoint video,'' in \emph{Proceedings of the IEEE/CVF Conference on Computer Vision and Pattern Recognition}, 2021, pp. 9421--9431.

\bibitem{li2021neural}
T.~Li, M.~Slavcheva, M.~Zollhoefer, S.~Green, C.~Lassner, C.~Kim, T.~Schmidt, S.~Lovegrove, M.~Goesele, and Z.~Lv, ``Neural {3D} video synthesis,'' 2021.

\bibitem{du2021neural}
Y.~Du, Y.~Zhang, H.-X. Yu, J.~B. Tenenbaum, and J.~Wu, ``Neural radiance flow for 4d view synthesis and video processing,'' in \emph{2021 IEEE/CVF International Conference on Computer Vision (ICCV)}.\hskip 1em plus 0.5em minus 0.4em\relax IEEE Computer Society, 2021, pp. 14\,304--14\,314.

\bibitem{wang2021neural}
C.~Wang, B.~Eckart, S.~Lucey, and O.~Gallo, ``Neural trajectory fields for dynamic novel view synthesis,'' \emph{arXiv preprint arXiv:2105.05994}, 2021.

\bibitem{gao2021dynamic}
C.~Gao, A.~Saraf, J.~Kopf, and J.-B. Huang, ``Dynamic view synthesis from dynamic monocular video,'' in \emph{Proceedings of the IEEE/CVF International Conference on Computer Vision}, 2021, pp. 5712--5721.

\bibitem{wang2022mixed}
F.~Wang, S.~Tan, X.~Li, Z.~Tian, and H.~Liu, ``Mixed neural voxels for fast multi-view video synthesis,'' \emph{arXiv preprint arXiv:2212.00190}, 2022.

\bibitem{song2022nerfplayer}
L.~Song, A.~Chen, Z.~Li, Z.~Chen, L.~Chen, J.~Yuan, Y.~Xu, and A.~Geiger, ``Nerfplayer: A streamable dynamic scene representation with decomposed neural radiance fields,'' \emph{arXiv preprint arXiv:2210.15947}, 2022.

\bibitem{park2021nerfies}
K.~Park, U.~Sinha, J.~T. Barron, S.~Bouaziz, D.~B. Goldman, S.~M. Seitz, and R.~Martin-Brualla, ``Nerfies: Deformable neural radiance fields,'' in \emph{Proceedings of the IEEE/CVF International Conference on Computer Vision}, 2021, pp. 5865--5874.

\bibitem{park2021hypernerf}
K.~Park, U.~Sinha, P.~Hedman, J.~T. Barron, S.~Bouaziz, D.~B. Goldman, R.~Martin-Brualla, and S.~M. Seitz, ``{HyperNeRF}: a higher-dimensional representation for topologically varying neural radiance fields,'' \emph{ACM Transactions on Graphics (TOG)}, vol.~40, no.~6, pp. 1--12, 2021.

\bibitem{li2022streaming}
L.~Li, Z.~Shen, Z.~Wang, L.~Shen, and P.~Tan, ``Streaming radiance fields for 3d video synthesis,'' \emph{arXiv preprint arXiv:2210.14831}, 2022.

\bibitem{shao2022Tensor4d}
R.~Shao, Z.~Zheng, H.~Tu, B.~Liu, H.~Zhang, and Y.~Liu, ``Tensor4d: Efficient neural 4d decomposition for high-fidelity dynamic reconstruction and rendering,'' \emph{arXiv preprint arXiv:2211.11610}, 2022.

\bibitem{jang2022d}
H.~Jang and D.~Kim, ``D-tensorf: Tensorial radiance fields for dynamic scenes,'' \emph{arXiv preprint arXiv:2212.02375}, 2022.

\bibitem{fang2022fast}
J.~Fang, T.~Yi, X.~Wang, L.~Xie, X.~Zhang, W.~Liu, M.~Nie{\ss}ner, and Q.~Tian, ``Fast dynamic radiance fields with time-aware neural voxels,'' in \emph{SIGGRAPH Asia 2022 Conference Papers}, 2022.

\bibitem{kappel2022fast}
M.~Kappel, V.~Golyanik, S.~Castillo, C.~Theobalt, and M.~Magnor, ``Fast non-rigid radiance fields from monocularized data,'' \emph{arXiv preprint arXiv:2212.01368}, 2022.

\bibitem{li2023instant3d}
J.~Li, H.~Tan, K.~Zhang, Z.~Xu, F.~Luan, Y.~Xu, Y.~Hong, K.~Sunkavalli, G.~Shakhnarovich, and S.~Bi, ``Instant3d: Fast text-to-3d with sparse-view generation and large reconstruction model,'' \emph{arXiv preprint arXiv:2311.06214}, 2023.

\bibitem{lin2023magic3d}
C.-H. Lin, J.~Gao, L.~Tang, T.~Takikawa, X.~Zeng, X.~Huang, K.~Kreis, S.~Fidler, M.-Y. Liu, and T.-Y. Lin, ``Magic3d: High-resolution text-to-3d content creation,'' in \emph{Proceedings of the IEEE/CVF Conference on Computer Vision and Pattern Recognition}, 2023, pp. 300--309.

\bibitem{wang2023score}
H.~Wang, X.~Du, J.~Li, R.~A. Yeh, and G.~Shakhnarovich, ``Score jacobian chaining: Lifting pretrained 2d diffusion models for 3d generation,'' in \emph{Proceedings of the IEEE/CVF Conference on Computer Vision and Pattern Recognition}, 2023, pp. 12\,619--12\,629.

\bibitem{zhu2023hifa}
J.~Zhu and P.~Zhuang, ``Hifa: High-fidelity text-to-3d with advanced diffusion guidance,'' \emph{arXiv preprint arXiv:2305.18766}, 2023.

\bibitem{liang2023luciddreamer}
Y.~Liang, X.~Yang, J.~Lin, H.~Li, X.~Xu, and Y.~Chen, ``Luciddreamer: Towards high-fidelity text-to-3d generation via interval score matching,'' \emph{arXiv preprint arXiv:2311.11284}, 2023.

\bibitem{Huang2023DreamTimeAI}
Y.~Huang, J.~Wang, Y.~Shi, X.~Qi, Z.~Zha, and L.~Zhang, ``Dreamtime: An improved optimization strategy for text-to-3d content creation,'' \emph{ArXiv}, vol. abs/2306.12422, 2023.

\bibitem{Raj2023DreamBooth3DST}
A.~Raj, S.~Kaza, B.~Poole, M.~Niemeyer, N.~Ruiz, B.~Mildenhall, S.~Zada, K.~Aberman, M.~Rubinstein, J.~T. Barron, Y.~Li, and V.~Jampani, ``Dreambooth3d: Subject-driven text-to-3d generation,'' \emph{ArXiv}, vol. abs/2303.13508, 2023.

\bibitem{Ruiz2022DreamBoothFT}
N.~Ruiz, Y.~Li, V.~Jampani, Y.~Pritch, M.~Rubinstein, and K.~Aberman, ``Dreambooth: Fine tuning text-to-image diffusion models for subject-driven generation,'' \emph{2023 IEEE/CVF Conference on Computer Vision and Pattern Recognition (CVPR)}, pp. 22\,500--22\,510, 2022.

\bibitem{metzer2023latent}
G.~Metzer, E.~Richardson, O.~Patashnik, R.~Giryes, and D.~Cohen-Or, ``Latent-nerf for shape-guided generation of 3d shapes and textures,'' in \emph{Proceedings of the IEEE/CVF Conference on Computer Vision and Pattern Recognition}, 2023, pp. 12\,663--12\,673.

\bibitem{xu2023dream3d}
J.~Xu, X.~Wang, W.~Cheng, Y.-P. Cao, Y.~Shan, X.~Qie, and S.~Gao, ``Dream3d: Zero-shot text-to-3d synthesis using 3d shape prior and text-to-image diffusion models,'' in \emph{Proceedings of the IEEE/CVF Conference on Computer Vision and Pattern Recognition}, 2023, pp. 20\,908--20\,918.

\bibitem{deng2023nerdi}
C.~Deng, C.~Jiang, C.~R. Qi, X.~Yan, Y.~Zhou, L.~Guibas, D.~Anguelov \emph{et~al.}, ``Nerdi: Single-view nerf synthesis with language-guided diffusion as general image priors,'' in \emph{Proceedings of the IEEE/CVF Conference on Computer Vision and Pattern Recognition}, 2023, pp. 20\,637--20\,647.

\bibitem{gu2023nerfdiff}
J.~Gu, A.~Trevithick, K.-E. Lin, J.~M. Susskind, C.~Theobalt, L.~Liu, and R.~Ramamoorthi, ``Nerfdiff: Single-image view synthesis with nerf-guided distillation from 3d-aware diffusion,'' in \emph{International Conference on Machine Learning}.\hskip 1em plus 0.5em minus 0.4em\relax PMLR, 2023, pp. 11\,808--11\,826.

\bibitem{qian2023magic123}
G.~Qian, J.~Mai, A.~Hamdi, J.~Ren, A.~Siarohin, B.~Li, H.-Y. Lee, I.~Skorokhodov, P.~Wonka, S.~Tulyakov \emph{et~al.}, ``Magic123: One image to high-quality 3d object generation using both 2d and 3d diffusion priors,'' \emph{arXiv preprint arXiv:2306.17843}, 2023.

\bibitem{tang2023make}
J.~Tang, T.~Wang, B.~Zhang, T.~Zhang, R.~Yi, L.~Ma, and D.~Chen, ``Make-it-3d: High-fidelity 3d creation from a single image with diffusion prior,'' \emph{arXiv preprint arXiv:2303.14184}, 2023.

\bibitem{ye2023consistent}
J.~Ye, P.~Wang, K.~Li, Y.~Shi, and H.~Wang, ``Consistent-1-to-3: Consistent image to 3d view synthesis via geometry-aware diffusion models,'' \emph{arXiv preprint arXiv:2310.03020}, 2023.

\bibitem{liu2023syncdreamer}
Y.~Liu, C.~Lin, Z.~Zeng, X.~Long, L.~Liu, T.~Komura, and W.~Wang, ``Syncdreamer: Generating multiview-consistent images from a single-view image,'' \emph{arXiv preprint arXiv:2309.03453}, 2023.

\bibitem{zeng2023ipdreamer}
B.~Zeng, S.~Li, Y.~Feng, H.~Li, S.~Gao, J.~Liu, H.~Li, X.~Tang, J.~Liu, and B.~Zhang, ``Ipdreamer: Appearance-controllable 3d object generation with image prompts,'' \emph{arXiv preprint arXiv:2310.05375}, 2023.

\bibitem{yu2023hifi}
W.~Yu, L.~Yuan, Y.-P. Cao, X.~Gao, X.~Li, L.~Quan, Y.~Shan, and Y.~Tian, ``Hifi-123: Towards high-fidelity one image to 3d content generation,'' \emph{arXiv preprint arXiv:2310.06744}, 2023.

\bibitem{shi2023zero123++}
R.~Shi, H.~Chen, Z.~Zhang, M.~Liu, C.~Xu, X.~Wei, L.~Chen, C.~Zeng, and H.~Su, ``Zero123++: a single image to consistent multi-view diffusion base model,'' \emph{arXiv preprint arXiv:2310.15110}, 2023.

\bibitem{weng2023consistent123}
H.~Weng, T.~Yang, J.~Wang, Y.~Li, T.~Zhang, C.~Chen, and L.~Zhang, ``Consistent123: Improve consistency for one image to 3d object synthesis,'' \emph{arXiv preprint arXiv:2310.08092}, 2023.

\bibitem{lin2023consistent123}
Y.~Lin, H.~Han, C.~Gong, Z.~Xu, Y.~Zhang, and X.~Li, ``Consistent123: One image to highly consistent 3d asset using case-aware diffusion priors,'' \emph{arXiv preprint arXiv:2309.17261}, 2023.

\bibitem{sun2023dreamcraft3d}
J.~Sun, B.~Zhang, R.~Shao, L.~Wang, W.~Liu, Z.~Xie, and Y.~Liu, ``Dreamcraft3d: Hierarchical 3d generation with bootstrapped diffusion prior,'' \emph{arXiv preprint arXiv:2310.16818}, 2023.

\bibitem{long2023wonder3d}
X.~Long, Y.-C. Guo, C.~Lin, Y.~Liu, Z.~Dou, L.~Liu, Y.~Ma, S.-H. Zhang, M.~Habermann, C.~Theobalt \emph{et~al.}, ``Wonder3d: Single image to 3d using cross-domain diffusion,'' \emph{arXiv preprint arXiv:2310.15008}, 2023.

\bibitem{Wang2023ImageDreamIM}
P.~Wang and Y.~Shi, ``Imagedream: Image-prompt multi-view diffusion for 3d generation,'' \emph{ArXiv}, vol. abs/2312.02201, 2023.

\bibitem{liu2023one}
M.~Liu, C.~Xu, H.~Jin, L.~Chen, Z.~Xu, H.~Su \emph{et~al.}, ``One-2-3-45: Any single image to 3d mesh in 45 seconds without per-shape optimization,'' \emph{arXiv preprint arXiv:2306.16928}, 2023.

\bibitem{liu2023one++}
M.~Liu, R.~Shi, L.~Chen, Z.~Zhang, C.~Xu, X.~Wei, H.~Chen, C.~Zeng, J.~Gu, and H.~Su, ``One-2-3-45++: Fast single image to 3d objects with consistent multi-view generation and 3d diffusion,'' \emph{arXiv preprint arXiv:2311.07885}, 2023.

\bibitem{hong2023lrm}
Y.~Hong, K.~Zhang, J.~Gu, S.~Bi, Y.~Zhou, D.~Liu, F.~Liu, K.~Sunkavalli, T.~Bui, and H.~Tan, ``Lrm: Large reconstruction model for single image to 3d,'' \emph{arXiv preprint arXiv:2311.04400}, 2023.

\bibitem{haque2023instruct}
A.~Haque, M.~Tancik, A.~A. Efros, A.~Holynski, and A.~Kanazawa, ``Instruct-nerf2nerf: Editing 3d scenes with instructions,'' \emph{arXiv preprint arXiv:2303.12789}, 2023.

\bibitem{sella2023vox}
E.~Sella, G.~Fiebelman, P.~Hedman, and H.~Averbuch-Elor, ``Vox-e: Text-guided voxel editing of 3d objects,'' in \emph{Proceedings of the IEEE/CVF International Conference on Computer Vision}, 2023, pp. 430--440.

\bibitem{zhuang2023dreameditor}
J.~Zhuang, C.~Wang, L.~Lin, L.~Liu, and G.~Li, ``Dreameditor: Text-driven 3d scene editing with neural fields,'' in \emph{SIGGRAPH Asia 2023 Conference Papers}, 2023, pp. 1--10.

\bibitem{li2023focaldreamer}
Y.~Li, Y.~Dou, Y.~Shi, Y.~Lei, X.~Chen, Y.~Zhang, P.~Zhou, and B.~Ni, ``Focaldreamer: Text-driven 3d editing via focal-fusion assembly,'' \emph{arXiv preprint arXiv:2308.10608}, 2023.

\bibitem{cheng2023progressive3d}
X.~Cheng, T.~Yang, J.~Wang, Y.~Li, L.~Zhang, J.~Zhang, and L.~Yuan, ``Progressive3d: Progressively local editing for text-to-3d content creation with complex semantic prompts,'' \emph{arXiv preprint arXiv:2310.11784}, 2023.

\bibitem{mirzaei2023watch}
A.~Mirzaei, T.~Aumentado-Armstrong, M.~A. Brubaker, J.~Kelly, A.~Levinshtein, K.~G. Derpanis, and I.~Gilitschenski, ``Watch your steps: Local image and scene editing by text instructions,'' \emph{arXiv preprint arXiv:2308.08947}, 2023.

\bibitem{fang2023text}
S.~Fang, Y.~Wang, Y.~Yang, Y.-H. Tsai, W.~Ding, M.-H. Yang, and S.~Zhou, ``Text-driven editing of 3d scenes without retraining,'' \emph{arXiv preprint arXiv:2309.04917}, 2023.

\bibitem{Tang2023DreamGaussianGG}
J.~Tang, J.~Ren, H.~Zhou, Z.~Liu, and G.~Zeng, ``Dreamgaussian: Generative gaussian splatting for efficient 3d content creation,'' \emph{ArXiv}, vol. abs/2309.16653, 2023.

\bibitem{Yi2023GaussianDreamerFG}
T.~Yi, J.~Fang, G.~Wu, L.~Xie, X.~Zhang, W.~Liu, Q.~Tian, and X.~Wang, ``Gaussiandreamer: Fast generation from text to 3d gaussian splatting with point cloud priors,'' \emph{ArXiv}, vol. abs/2310.08529, 2023.

\bibitem{Chen2023Textto3DUG}
Z.~Chen, F.~Wang, and H.~Liu, ``Text-to-3d using gaussian splatting,'' \emph{ArXiv}, vol. abs/2309.16585, 2023.

\bibitem{li2023generative}
C.~Li, C.~Zhang, A.~Waghwase, L.-H. Lee, F.~Rameau, Y.~Yang, S.-H. Bae, and C.~S. Hong, ``Generative ai meets 3d: A survey on text-to-3d in aigc era,'' \emph{arXiv preprint arXiv:2305.06131}, 2023.

\bibitem{zheng2023unified}
Y.~Zheng, X.~Li, K.~Nagano, S.~Liu, O.~Hilliges, and S.~De~Mello, ``A unified approach for text-and image-guided 4d scene generation,'' \emph{arXiv preprint arXiv:2311.16854}, 2023.

\bibitem{bahmani20234d}
S.~Bahmani, I.~Skorokhodov, V.~Rong, G.~Wetzstein, L.~Guibas, P.~Wonka, S.~Tulyakov, J.~J. Park, A.~Tagliasacchi, and D.~B. Lindell, ``4d-fy: Text-to-4d generation using hybrid score distillation sampling,'' \emph{arXiv preprint arXiv:2311.17984}, 2023.

\bibitem{Luo2023VideoFusionDD}
\BIBentryALTinterwordspacing
Z.~Luo, D.~Chen, Y.~Zhang, Y.~Huang, L.~Wang, Y.~Shen, D.~Zhao, J.~Zhou, and T.-P. Tan, ``Videofusion: Decomposed diffusion models for high-quality video generation,'' in \emph{2023 IEEE/CVF Conference on Computer Vision and Pattern Recognition (CVPR)}, 2023, pp. 10\,209--10\,218. [Online]. Available: \url{https://api.semanticscholar.org/CorpusID:257532642}
\BIBentrySTDinterwordspacing

\bibitem{Ling2023AlignYG}
H.~Ling, S.~W. Kim, A.~Torralba, S.~Fidler, and K.~Kreis, ``Align your gaussians: Text-to-4d with dynamic 3d gaussians and composed diffusion models,'' \emph{arXiv preprint}, 2023.

\bibitem{bahmani2024tc4d}
S.~Bahmani, X.~Liu, Y.~Wang, I.~Skorokhodov, V.~Rong, Z.~Liu, X.~Liu, J.~J. Park, S.~Tulyakov, G.~Wetzstein \emph{et~al.}, ``Tc4d: Trajectory-conditioned text-to-4d generation,'' \emph{arXiv preprint arXiv:2403.17920}, 2024.

\bibitem{xu2024comp4d}
D.~Xu, H.~Liang, N.~P. Bhatt, H.~Hu, H.~Liang, K.~N. Plataniotis, and Z.~Wang, ``Comp4d: Llm-guided compositional 4d scene generation,'' \emph{arXiv preprint arXiv:2403.16993}, 2024.

\bibitem{wang2024vidu4d}
Y.~Wang, X.~Wang, Z.~Chen, Z.~Wang, F.~Sun, and J.~Zhu, ``Vidu4d: Single generated video to high-fidelity 4d reconstruction with dynamic gaussian surfels,'' \emph{arXiv preprint arXiv:2405.16822}, 2024.

\bibitem{liang2024diffusion4d}
H.~Liang, Y.~Yin, D.~Xu, H.~Liang, Z.~Wang, K.~N. Plataniotis, Y.~Zhao, and Y.~Wei, ``Diffusion4d: Fast spatial-temporal consistent 4d generation via video diffusion models,'' \emph{arXiv preprint arXiv:2405.16645}, 2024.

\bibitem{sun2024eg4d}
Q.~Sun, Z.~Guo, Z.~Wan, J.~N. Yan, S.~Yin, W.~Zhou, J.~Liao, and H.~Li, ``Eg4d: Explicit generation of 4d object without score distillation,'' \emph{arXiv preprint arXiv:2405.18132}, 2024.

\bibitem{wu20244d}
G.~Wu, T.~Yi, J.~Fang, L.~Xie, X.~Zhang, W.~Wei, W.~Liu, Q.~Tian, and X.~Wang, ``4d gaussian splatting for real-time dynamic scene rendering,'' in \emph{Proceedings of the IEEE/CVF Conference on Computer Vision and Pattern Recognition}, 2024, pp. 20\,310--20\,320.

\bibitem{miao2024pla4d}
Q.~Miao, Y.~Luo, and Y.~Yang, ``Pla4d: Pixel-level alignments for text-to-4d gaussian splatting,'' \emph{arXiv preprint arXiv:2405.19957}, 2024.

\bibitem{fu2024sync4d}
Z.~Fu, J.~Wei, W.~Shen, C.~Song, X.~Yang, F.~Liu, X.~Yang, and G.~Lin, ``Sync4d: Video guided controllable dynamics for physics-based 4d generation,'' \emph{arXiv preprint arXiv:2405.16849}, 2024.

\bibitem{zhang2024physdreamer}
T.~Zhang, H.-X. Yu, R.~Wu, B.~Y. Feng, C.~Zheng, N.~Snavely, J.~Wu, and W.~T. Freeman, ``Physdreamer: Physics-based interaction with 3d objects via video generation,'' \emph{arXiv preprint arXiv:2404.13026}, 2024.

\bibitem{jiang2023consistent4d}
Y.~Jiang, L.~Zhang, J.~Gao, W.~Hu, and Y.~Yao, ``Consistent4d: Consistent 360 $\{$$\backslash$deg$\}$ dynamic object generation from monocular video,'' \emph{arXiv preprint arXiv:2311.02848}, 2023.

\bibitem{yin20234dgen}
Y.~Yin, D.~Xu, Z.~Wang, Y.~Zhao, and Y.~Wei, ``4dgen: Grounded 4d content generation with spatial-temporal consistency,'' \emph{arXiv preprint arXiv:2312.17225}, 2023.

\bibitem{zhao2023animate124}
Y.~Zhao, Z.~Yan, E.~Xie, L.~Hong, Z.~Li, and G.~H. Lee, ``Animate124: Animating one image to 4d dynamic scene,'' \emph{arXiv preprint arXiv:2311.14603}, 2023.

\bibitem{Wang2023AnimatableDreamerTN}
X.~Wang, Y.~Wang, J.~Ye, Z.~Wang, F.~Sun, P.~Liu, L.~Wang, K.~Sun, X.~Wang, and B.~He, ``Animatabledreamer: Text-guided non-rigid 3d model generation and reconstruction with canonical score distillation,'' \emph{arXiv preprint}, 2023.

\bibitem{zhang2024magicpose4d}
H.~Zhang, D.~Chang, F.~Li, M.~Soleymani, and N.~Ahuja, ``Magicpose4d: Crafting articulated models with appearance and motion control,'' \emph{arXiv preprint arXiv:2405.14017}, 2024.

\bibitem{wu2024sc4d}
Z.~Wu, C.~Yu, Y.~Jiang, C.~Cao, F.~Wang, and X.~Bai, ``Sc4d: Sparse-controlled video-to-4d generation and motion transfer,'' \emph{arXiv preprint arXiv:2404.03736}, 2024.

\bibitem{huang2023sc}
Y.-H. Huang, Y.-T. Sun, Z.~Yang, X.~Lyu, Y.-P. Cao, and X.~Qi, ``Sc-gs: Sparse-controlled gaussian splatting for editable dynamic scenes,'' \emph{arXiv preprint arXiv:2312.14937}, 2023.

\bibitem{chu2024dreamscene4d}
W.-H. Chu, L.~Ke, and K.~Fragkiadaki, ``Dreamscene4d: Dynamic multi-object scene generation from monocular videos,'' \emph{arXiv preprint arXiv:2405.02280}, 2024.

\bibitem{zeng2024stag4d}
Y.~Zeng, Y.~Jiang, S.~Zhu, Y.~Lu, Y.~Lin, H.~Zhu, W.~Hu, X.~Cao, and Y.~Yao, ``Stag4d: Spatial-temporal anchored generative 4d gaussians,'' \emph{arXiv preprint arXiv:2403.14939}, 2024.

\bibitem{zhang20244diffusion}
H.~Zhang, X.~Chen, Y.~Wang, X.~Liu, Y.~Wang, and Y.~Qiao, ``4diffusion: Multi-view video diffusion model for 4d generation,'' \emph{arXiv preprint arXiv:2405.20674}, 2024.

\bibitem{kajiya1984ray}
J.~T. Kajiya and B.~P. Von~Herzen, ``Ray tracing volume densities,'' \emph{ACM SIGGRAPH Computer Graphics}, vol.~18, no.~3, pp. 165--174, 1984.

\bibitem{wang2023videofactory}
W.~Wang, H.~Yang, Z.~Tuo, H.~He, J.~Zhu, J.~Fu, and J.~Liu, ``Videofactory: Swap attention in spatiotemporal diffusions for text-to-video generation,'' \emph{arXiv preprint arXiv:2305.10874}, 2023.

\bibitem{hu2021lora}
E.~J. Hu, Y.~Shen, P.~Wallis, Z.~Allen-Zhu, Y.~Li, S.~Wang, L.~Wang, and W.~Chen, ``Lora: Low-rank adaptation of large language models,'' \emph{arXiv preprint arXiv:2106.09685}, 2021.

\bibitem{radford2021learning}
A.~Radford, J.~W. Kim, C.~Hallacy, A.~Ramesh, G.~Goh, S.~Agarwal, G.~Sastry, A.~Askell, P.~Mishkin, J.~Clark \emph{et~al.}, ``Learning transferable visual models from natural language supervision,'' in \emph{International Conference on Machine Learning}.\hskip 1em plus 0.5em minus 0.4em\relax PMLR, 2021, pp. 8748--8763.

\bibitem{li2022blip}
J.~Li, D.~Li, C.~Xiong, and S.~Hoi, ``Blip: Bootstrapping language-image pre-training for unified vision-language understanding and generation,'' in \emph{International Conference on Machine Learning}.\hskip 1em plus 0.5em minus 0.4em\relax PMLR, 2022, pp. 12\,888--12\,900.

\end{thebibliography}
}

\end{document}